\documentclass{article}

\usepackage{microtype}
\usepackage{graphicx}
\usepackage{subfigure}
\usepackage{booktabs} 

\usepackage{hyperref}
\usepackage{float}
\usepackage{chapterbib}

\usepackage[accepted]{icml2025}

\usepackage{amsthm}

\usepackage[utf8]{inputenc} 
\usepackage[T1]{fontenc}    
\usepackage{hyperref}       
\usepackage{url}            
\usepackage{booktabs}       
\usepackage{amsfonts}       
\usepackage{nicefrac}       
\usepackage{microtype}      
\usepackage{xcolor}         
\usepackage{graphicx}
\usepackage{float}
\usepackage{wrapfig}
\usepackage{subfigure}
\usepackage{amsmath}
\usepackage{amssymb}
\usepackage{multirow}
\usepackage{diagbox}
\usepackage{colortbl}
\usepackage{tabularx}
\usepackage{adjustbox}
\usepackage{pifont}
\usepackage{bm}
\usepackage{mathtools}
\usepackage{algpseudocode}
\usepackage{listings}
\usepackage[capitalize,noabbrev]{cleveref}

\definecolor{rebuttal}{RGB}{0,0,0}

\newcommand{\reb}[1]{\textcolor{rebuttal}{#1}}

\theoremstyle{plain}

\theoremstyle{definition}

\theoremstyle{remark}

\usepackage[textsize=tiny]{todonotes}

\icmltitlerunning{Lumina-mGPT: Illuminate Flexible Photorealistic Text-to-Image Generation with Multimodal Generative Pretraining}

\begin{document}

\twocolumn[
\icmltitle{Lumina-mGPT: Illuminate Flexible Photorealistic Text-to-Image Generation with Multimodal Generative Pretraining}



\icmlsetsymbol{equal}{$\textbf{*}$} 
\icmlsetsymbol{corresp}{$\P$} 
\icmlsetsymbol{lead}{$\blacklozenge$} 
\icmlsetsymbol{sep}{\!\!,\!}

\begin{icmlauthorlist}
\icmlauthor{Dongyang Liu}{equal,CUHK,AILAB}
\icmlauthor{Shitian Zhao}{equal,AILAB}
\icmlauthor{Le Zhuo}{equal,CUHK,AILAB}
\icmlauthor{Weifeng Lin}{equal,CUHK}
\icmlauthor{Yi Xin}{AILAB}
\icmlauthor{Xinyue Li}{AILAB}
\icmlauthor{Qi Qin}{AILAB}
\icmlauthor{Yu Qiao}{AILAB}
\icmlauthor{Hongsheng Li}{corresp,CUHK}
\icmlauthor{Peng Gao}{corresp,lead,AILAB}
\end{icmlauthorlist}


\icmlaffiliation{CUHK}{The Chinese University of Hong Kong}
\icmlaffiliation{AILAB}{Shanghai AI Laboratory}

\icmlcorrespondingauthor{Hongsheng Li}{hsli@ee.cuhk.edu.hk}
\icmlcorrespondingauthor{Peng Gao}{gaopeng@pjlab.org.cn}

\icmlkeywords{Machine Learning, ICML}

\vskip 0.3in
]



\printAffiliationsAndNotice{\icmlEqualContribution} 

\begin{abstract}
We present Lumina-mGPT, a family of multimodal autoregressive models capable of various vision and language tasks, particularly excelling in generating flexible photorealistic images from text descriptions. \reb{By initializing from \textit{\textbf{m}ultimodal \textbf{G}enerative \textbf{P}re\textbf{T}raining} (mGPT), we demonstrate that decoder-only Autoregressive (AR) model can achieve image generation performance comparable to modern diffusion model with high efficiency through \textit{Flexible Progressive Supervised Finetuning} (FP-SFT). Equipped with our proposed \textit{\textbf{Un}ambiguous \textbf{i}mage \textbf{Rep}resentation} (Uni-Rep), Lumina-mGPT can flexibly generate high-quality images of varying aspect ratios. Building on the strong image generation capabilities, we further explore \textit{Ominiponent Supervised Finetuning} (Omni-SFT), an initial attempt to elevate Lumina-mGPT into a unified multi-modal generalist}. The resulting model demonstrates versatile multimodal capabilities, including visual generation tasks like text-to-image/multiview generation and controllable generation, visual recognition tasks like segmentation and depth estimation, and vision-language tasks like multi-turn visual question answering, showing the rosy potential of the technical direction. Codes and checkpoints are available at \url{https://github.com/Alpha-VLLM/Lumina-mGPT}.
\end{abstract}

\begin{figure*}
    \centering
    \includegraphics[width=0.85\linewidth]{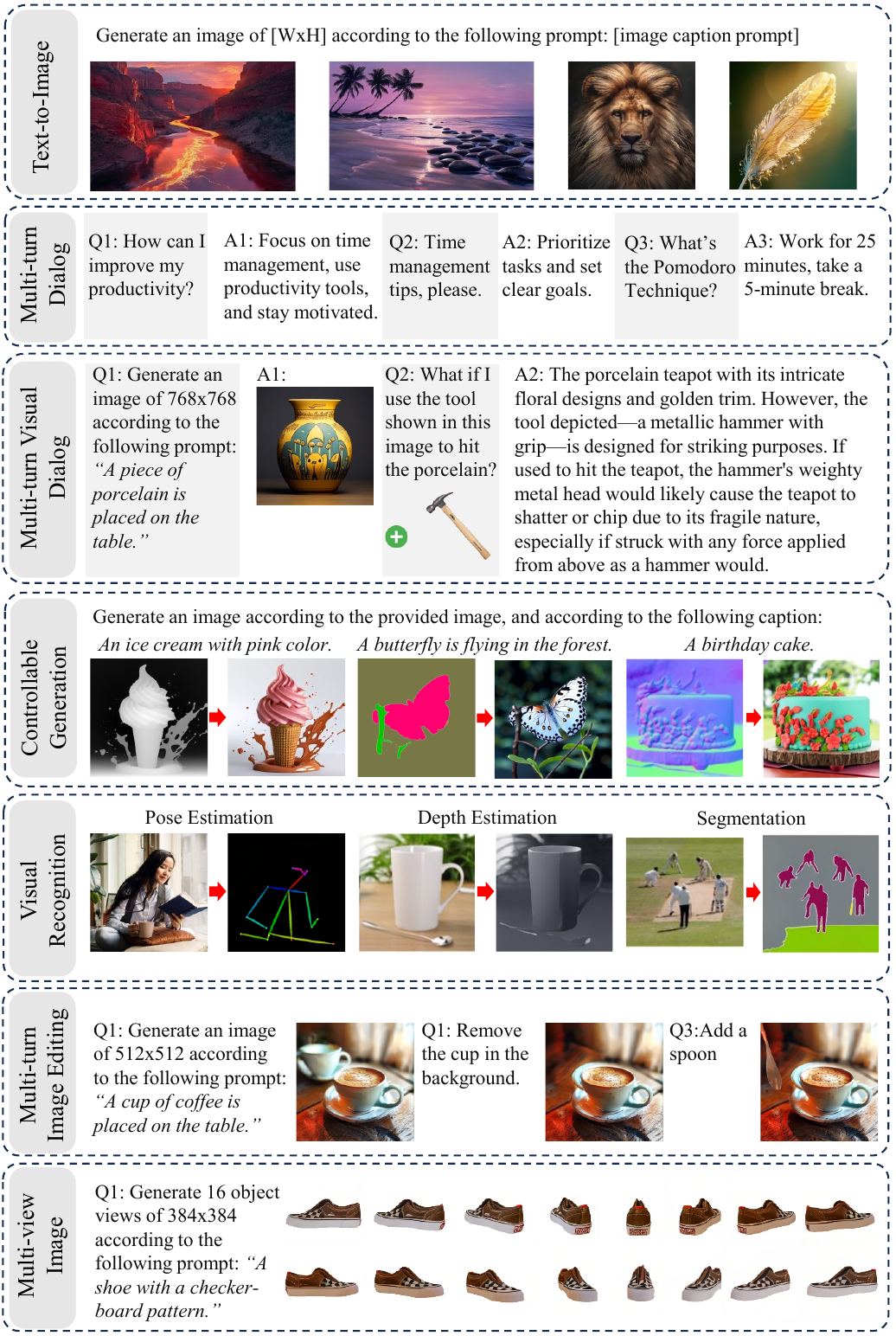}
    \caption{Unified in the next-token prediction framework, Lumina-mGPT can perform a wide range of multi-modal tasks. See Figure~\ref{fig:demos} to Figure~\ref{fig:demos_all_p4} in the Appendix for more demonstrations.}
    \label{fig:all_demos}
\end{figure*}

\section{Introduction}
\label{sec:intro}
Seminal models, including DALL-E 3~\citep{dalle3}, Stable Diffusion 3~\citep{sd3}, and SoRA~\citep{sora}, have demonstrated superior performance in photorealistic image and video generation using diffusion-based generative modeling over continuous latent image features. In contrast, autoregressive (AR) generative models, which rely on ``next-token prediction'', have revolutionized text generation with groundbreaking reasoning abilities, as exemplified by models like GPT-4~\citep{gpt4} and Deepseek-v3~\citep{gemini}. 

However, AR-based generative modeling with vector-quantized image features remains a relatively less mature topic. While previous autoregressive efforts, such as DALL-E~\citep{dalle}, CogView~\citep{cogview}, Parti~\citep{parti}, OFA~\citep{ofa}, Unified-IO~\citep{unified-io,unified-io2}, LlamaGen~\citep{llamagen}, and Chameleon~\citep{chameleon}, have made important explorations, each of these methods suffers from one or more of the following limitations: (1) their image generation capabilities have been either unsatisfactory or constrained to academic benchmarks such as ImageNet~\citep{imagenet}; (2) they rely on complex architectures, such as encoder-decoder frameworks, which impede scalability and limit generalization to other tasks; (3) they are restricted in decoding resolution and flexibility, often producing small images with fixed resolutions; and (4) they lack task extensibility, making them unsuitable for broader scenarios that require integration with tasks like dense labeling or controllable image generation. A more detailed discussion of these challenges is provided in Appendix~\ref{app:limit_exist}.

\begin{table*}[t]
\centering\small
\caption{Design choices and capabilities of multimodal autoregressive approaches. Lumina-mGPT is the only model capable of both flexible photorealistic image generation and multimodal task unification, due to its decoder-only transformer design and multimodal generative pretraining.}
\vspace{0.5em}
\label{table:overview}
\begin{tabular}{@{}lccccc@{}}
\toprule
\multirow{2}{*}{Method} 
& Model & Multimodal & Photorealistic & Flexible Image & Task \\ 
& Architecture & Pretraining & Image Generation & Aspect Ratio & Extensibility \\
\midrule
DALL-E~\citep{dalle} & Decoder-only & \texttimes  & \texttimes & \texttimes & \texttimes \\ 
Cogview~\citep{cogview} & Decoder-only & \texttimes  & \texttimes & \texttimes & \texttimes \\
Parti~\citep{parti} & Encoder-Decoder & \texttimes  & \texttimes & \texttimes & \texttimes \\
LlamaGen~\citep{llamagen} & Encoder-Decoder & \texttimes & \texttimes & \texttimes & \texttimes \\
OFA~\citep{ofa} & Encoder-Decoder & \checkmark  & \texttimes & \texttimes & \checkmark \\
Unified-IO~\citep{unified-io} & Encoder-Decoder & \texttimes  & \texttimes & \texttimes & \checkmark \\
Unified-IO 2~\citep{unified-io2} & Encoder-Decoder & \checkmark  & \texttimes & \texttimes & \checkmark \\
CM3Leon~\citep{cm3leon} & Decoder-only & \checkmark  & \texttimes & \texttimes & \checkmark \\
Chameleon~\citep{chameleon} & Decoder-only & \checkmark  & \texttimes & \texttimes & \texttimes \\
\bf Lumina-mGPT & Decoder-only & \checkmark & \checkmark & \checkmark & \checkmark \\
\bottomrule
\end{tabular}
\end{table*}

To address the aforementioned challenges, we present Lumina-mGPT, a decoder-only transformer initiated with effective \textit{\textbf{m}ultimodal \textbf{G}enerative \textbf{P}re\textbf{T}raining} (mGPT) and then supervised-finetuned over flexible, high-quality, high-resolution discrete image tokens in a progressive manner. This framework illuminates flexible high-resolution photorealistic image generation and can be easily extended to solve various downstream tasks in a unified manner. We provide a detailed comparison of the architecture design and model capabilities of existing multimodal autoregressive approaches in Table~\ref{table:overview}. The key features of Lumina-mGPT are outlined below: 

\noindent \underline{\textbf{\textit{\ding{192} Simple Decoder-Only Architecture:}}} We adopt a simple decoder-only architecture, which offers a significant advantage over more complex designs such as encoder-decoder architectures. Decoder-only models provide an elegant and extensible framework that unifies various understanding and generation tasks across different modalities. This makes them a compelling choice for achieving true unification. Furthermore, by leveraging the same architecture as the rapidly evolving text-only LLMs, Lumina-mGPT can take advantage of well-established theories and infrastructure in the LLM community, including scaling properties~\citep{palm,gpt3} and advanced techniques for optimizing training and inference~\citep{flashattention,flashattention2,vllm}.

\noindent \underline{\textbf{\textit{\ding{193} Flexible High-quality Image Generation:}}} Despite the aforementioned strengths of decoder-only architecture, the image generation capabilities of such models remain limited, creating a gap between the potential and the reality of this architecture. We thus propose Flexible Progressive Supervised Finetuning (FP-SFT) to fully fulfill the potential of high-quality text-to-image generation. This approach starts with low-resolution discrete tokens and progressively transitions to high-resolution discrete tokens. Combined with \textit{\textbf{Un}ambiguous \textbf{i}mage \textbf{Rep}resentation} (Uni-Rep), this weak-to-strong SFT strategy effectively grants the model with the ability to generate high-quality photorealistic images with flexible aspect ratios. 

\noindent \underline{\textbf{\textit{\ding{194} Omnipotent Task Unification:}}} The high-quality image generation capabilities achieved during the FP-SFT stage provide the prerequisites for further exploring the fundamental strength, namely task unification, of the decoder-only architecture. We thus propose Omnipotent Supervised Finetuning (Omni-SFT), a preliminary attempt to create a multi-modal omnipotent generalist. Omni-SFT treats various tasks--such as multi-turn dialog, visual multi-turn understanding, dense labeling, text-to-image generation, text-to-multiview generation, image editing, and spatial-conditional image generation--as a unified discrete modeling task, allowing the model to achieve omnipotence via a unified natural language interface. 

\noindent \underline{\textbf{\textit{\ding{195} Effective Multimodal Generative Pretraining:}}} Instead of random initialization, Lumina-mGPT leverages an effective \textbf{m}ultimodal \textbf{G}enerative \textbf{P}ret\textbf{T}raining (mGPT). This representation is derived from a multimodal transformer trained at scale using a straightforward "next-token prediction" loss. The rich knowledge encapsulated in this pretraining significantly accelerates the learning process for text-to-image generation and the downstream vision-language tasks. Given the constraints in training resources, we adopt the mGPT representation directly from the pretrained Chameleon 7B and 30B models released by Meta~\citep{chameleon}.

\reb{The summarized contributions are as follows: (1) We are the first, particularly in the open-source domain, to demonstrate that a decoder-only AR model can achieve image generation performance on par with modern diffusion models. Besides, with mGPT initialization, this capability can be achieved with remarkable efficiency, requiring only 32 A100 GPUs over 7 days to train a 7B model. (2) We introduce UniRep, a novel image representation that enables decoder-only AR models to generate images with varying aspect ratios, offering greater flexibility in image generation. (3) Leveraging the strong image generation capabilities achieved, we propose Omni-SFT, a pioneering exploration aimed at elevating the model into a unified generalist. (4) We release the full-pipeline code implementation to facilitate further advancements from the research community.}

\begin{figure*}[t]
  \centering
  \includegraphics[width = \linewidth]{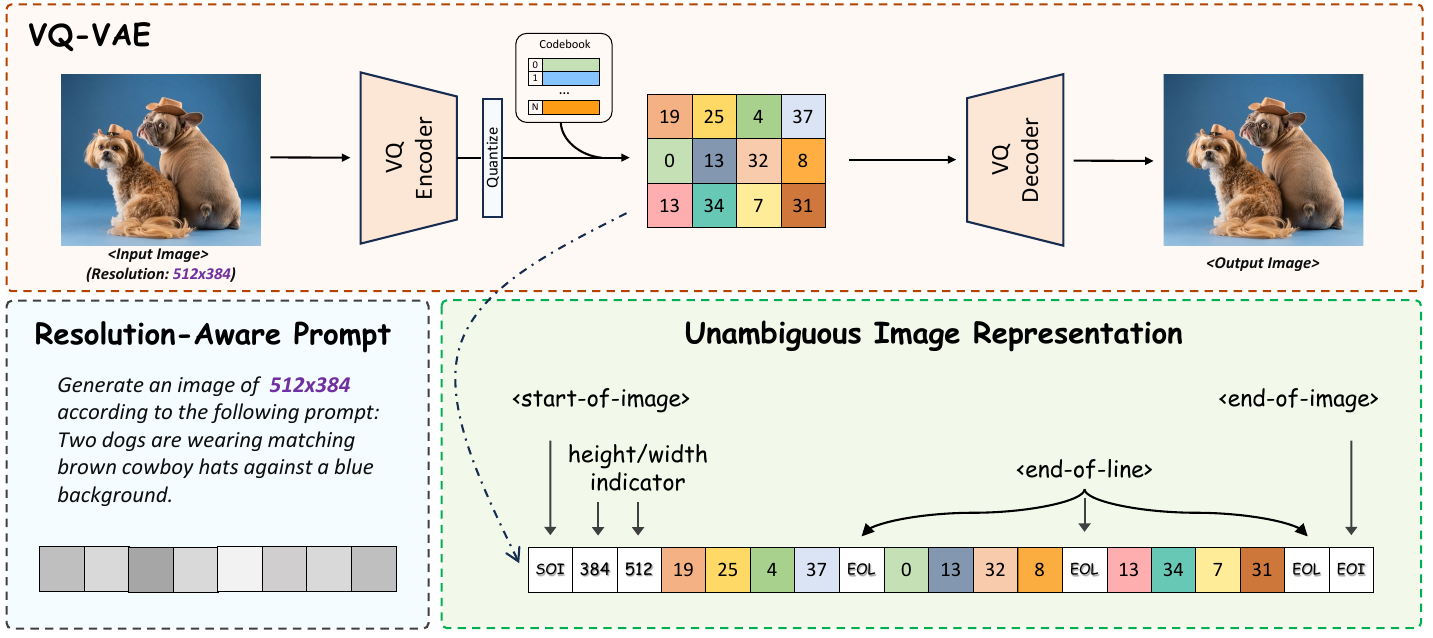}
  \vspace{-2em}
  \caption{Illustration of Resolution-Aware Prompt (bottom left) and Unambiguous Image Representation (bottom right). These designs are used in all supervised finetuning stages to eliminate the ambiguity in image representation, enabling flexible resolution image modeling.}
  \label{fig:prompt_representation}
\end{figure*}

\section{Methodolgy}
\label{sec:method}

Lumina-mGPT is a decoder-only transformer initialized with multimodal Generative PreTraining (mGPT) and finetuned over high-quality multimodal tokens derived from various tasks. Based on the robust mGPT representation and our proposed supervised finetuning strategies with unambiguous image representation, Lumina-mGPT achieves superior performance in photorealistic image generation and omnipotent task unification with high flexibility in image resolution and aspect ratio.

\subsection{Revisiting mGPT with Chameleon}
mGPT represents the family of models utilizing a decoder-only transformer architecture and pretrained on extensive multimodal token sequences.
We use the recent open-source model, Chameleon~\citep{chameleon}, as an example to illustrate the design choices and implementing details of mGPT.

\paragraph{Multimodal Tokenization}
To unify text and images into a multimodal token sequence, it is essential first to tokenize both text and image into discrete space. The choice of image tokenizer is especially crucial as it determines the upper limit of generation quality. Chameleon trains a byte pair encoding (BPE) tokenizer for text; for images, it adopts the quantization-based tokenization method following prior works~\citep{vqgan,parti,dalle}, converting continuous image patches into discrete tokens from a fixed codebook while reducing spatial dimensions. The quantized image tokens are then flattened into 1D sequence and concatenated with text tokens  to form a multimodal token sequence for unified modeling. 

\paragraph{Decoder-Only Transformer}
In contrast to the encoder-decoder architecture adopted by Unified-IO and Parti, the decoder-only architecture leads to a simpler and more extensible approach for multimodal generative modeling. By transforming text and image inputs into a unified sequence of discrete tokens, the understanding and generation within and across modalities can be elegantly unified, forming a rosy direction towards general intelligence. During training, mGPT models the conditional probability $p(x_t|x_1, ..., x_{t-1})$ of multimodal sequences using the standard next-token prediction objective. Additionally, Chameleon applies z-loss~\citep{palm} to stabilize the training. 

\paragraph{Limitations of Chameleon}
Although Chameleon demonstrates potential for joint image and text understanding within one decoder-only transformer, its image generation ability remains inferior to state-of-the-art diffusion-based frameworks~\citep{sd3,pixart-sigma,hunyuandit,kolors,lumina-next} in both quality and resolution flexibility. Moreover, it is worth noting that the image generation ability is even absent in the open-source version of Chameleon. Additionally, the capabilities of Chameleon are confined to vision-language and text-only tasks, excluding a broader range of vision-centric tasks. These include classic visual recognition tasks such as segmentation and depth prediction, as well as creative visual generation tasks like controllable generation and image editing. Lumina-mGPT is built upon Chameleon to unlock its full potential for flexible photorealistic image generation and to become a versatile vision generalist.

\subsection{Lumina-mGPT}
\subsubsection{Effective Initlization}
Large-scale pre-training and scalable model architecture have been widely verified as the golden path to advanced intelligence. As mGPTs like Chameleon are pretrained on large-scale datasets and have developed effective and generalizable representations, they are well-suited to serve as the starting point for flexible photorealistic image generation and beyond. In this work, we initialize from Chameleon with parameters ranging from 7B to 30B, using just 10M high-quality image-text data points.

\subsubsection{Supervised Finetuning for Lumina-mGPT}
\paragraph{Unambiguous Image Representation}
Existing methods represent images as 1D flattened sequences of 2D discrete image codes. While adequate for fixed resolutions, this approach becomes ambiguous when supporting variable resolutions, as with Lumina-mGPT. For instance, images with resolutions of 512 $\times$ 512, 256 $\times$ 1024, and 1024 $\times$ 256 can all be encoded into the same number of tokens, making it impossible to infer the original shape without examining the token contents. This ambiguity poses significant challenges for both image perception and generation.

To address this problem, we propose \textbf{Un}ambiguous \textbf{i}mage \textbf{Rep}resentation (Uni-Rep), which augments the image representations by adding extra height/width indicator tokens immediately after the \texttt{<start-of-image>} token and inserting \texttt{<end-of-line>} tokens after image tokens belonging to the same row. As shown in Figure~\ref{fig:prompt_representation}, this modification ensures that the original shape of the images can be accurately parsed from the 1D representation without additional context or delving into the contents of the image tokens. This enhancement provides the foundation for Lumina-mGPT's ability to perform image-related tasks at any resolution and aspect ratio. 

\textit{Discussions:} While either the height/width indicators or the \texttt{<end-of-line>} tokens can independently achieve disambiguation, we use both simultaneously due to their distinct benefits. When generating images, the height/width indicators, generated before any image tokens, pre-determine the shape of the image, aiding Lumina-mGPT in composing the image contents. The \texttt{<end-of-line>} tokens, on the other hand, serve as anchors, offering the 1D token sequence with additional explicit spatial information. Section~\ref{sec:attn_vis} shows empirical analysis over these tokens. 

\paragraph{Flexible Progressive Supervised Finetuning (FP-SFT)}
The FP-SFT process equips the pretrained mGPT with the capability to generate high-resolution images with flexible aspect ratios in a progressive manner. The process is divided into three stages, where the product of width and height approximates $512^2$, $768^2$, and $1024^2$, respectively. In each stage, a set of candidate resolutions with similar areas but different height-width ratios are prepared, and each image is matched to the most suitable resolution. In the low-resolution stage, shorter sequence lengths and the resulting high training throughput allow the model to quickly traverse a large amount of data, learning the general composition of images and a broad spectrum of visual concepts. Conversely, in the high-resolution stage, the model is expected to focus on learning high-frequency fine-grained details unique to high-resolution images. Benefiting from the strong foundation built during the high-throughput pretraining and low-resolution finetuning stages, the low-throughput high-resolution finetuning stage is data-efficient, thereby enhancing the overall efficiency of the FP-SFT process.

A meticulously curated dataset of high-resolution photorealistic image-text pairs is used for FP-SFT. Moreover, the pure-text data from OpenHermess~\citep{OpenHermes2.5} and the image-to-text data from Mini-Gemini~\citep{mgm} are also incorporated during training to prevent catastrophic forgetting. To provide users with a natural way to specify the desired resolution, we develop a resolution-aware prompt (Figure~\ref{fig:prompt_representation}). For each image and its corresponding description, the prompt is structured as follows:\newline
\texttt{
\small
Generate an image of \{width\}x\{height\} according to the following prompt:
\textbackslash{}n
\{description\}
}
\paragraph{Omnipotent Supervised Finetuning (Omni-SFT)}
While flexible photorealistic image generation is the primary target of Lumina-mGPT, we find that the resulting model after FP-SFT can be efficiently transferred to a wide spectrum of image understanding and generation tasks. We thus present Omni-SFT, a preliminary exploration toward boosting Lumina-mGPT to a visual generalist. Training tasks and data for Omni-SFT consists of the following: 

1. Single- and multi-turn language-guided image-editing with data from MagicBrush~\citep{zhang2024magicbrush} and SEED~\citep{SeedDataEdit} (only involving the real-world and multi-turn subsets).

2. Dense prediction tasks, including surface norm estimation from NYUv2~\cite{silberman2012indoor} and ScanNet~\cite{dai2017scannet}, depth estimation from Kitti v2~\cite{cabon2020virtual} and Sintel~\cite{butler2012naturalistic}, pose estimation from MSCOCO~\cite{lin2014microsoft}, semantic segmentation data annotated with OneFormer~\citep{jain2023oneformer} on image from Laion~\citep{schuhmann2022laion}, and grounding data from RefCOCO~\citep{kazemzadeh2014referitgame}.

3. In-house spatial-conditional image generation following ControlNet~\citep{controlnet}, with conditions including surface norm, depth, pose, and segmentation.

4. Text-conditional multiview generation using an internal dataset consisting of ~100k high-quality samples with rendered $384^2$ images from 16 viewpoints uniformly distributed in azimuth angles.

5. A small fraction of data sampled from those used in the previous FP-SFT process, including both text modeling and text-to-image generation to maintain its learned capabilities.

We tokenize all text and images into discrete tokens and formulate these tasks as a unified next-token prediction objective. Notably, we also incorporate tasks multiview generation, which requires generating a sequence of image frames, as a preliminary for video generation. As demonstrated in Section~\ref{sec:exp_omni_task_uni}, after Omni-SFT, Lumina-mGPT exhibits a general capability for completing a wide range of tasks other than text-to-image generation, indicating the potential for building a multimodal generalist along this direction.

\begin{figure}[t]
  \centering
  \includegraphics[width = \linewidth]{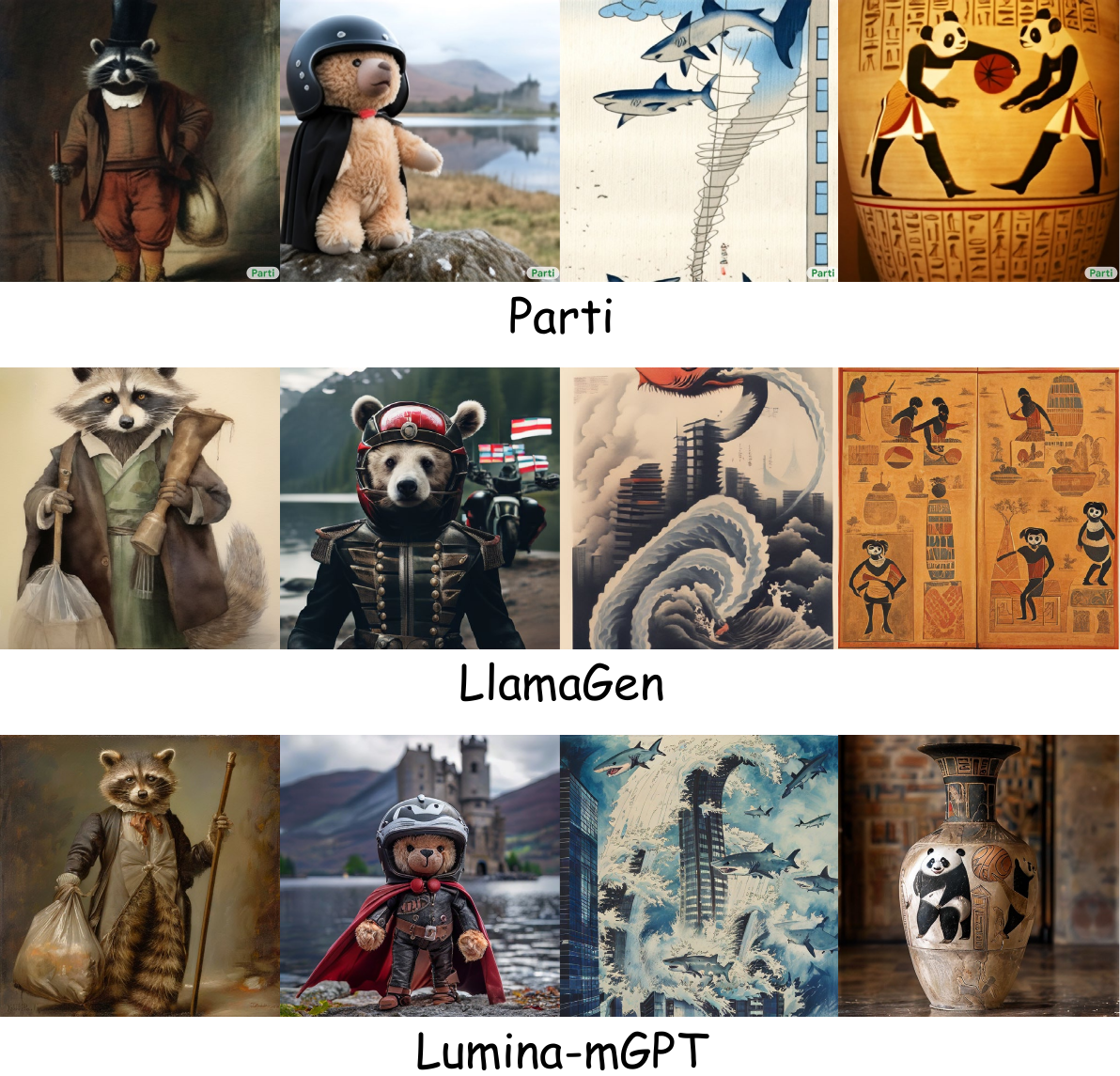}
  \vspace{-2em}
  \caption{Qualitative comparison on text-to-image generation. Prompts: (1) A raccoon wearing formal clothes, wearing a tophap and holding a cane. The raccoon is holding a garbage bag. Oil painting in the style of Rembrandt. (2) A teddy bear wearing a motorcycle helmet and cape is standing in front of Loch Awe with Kilchurn Castle behind him, dslr photo. (3) A tornado made of sharks crashing into a skyscraper. Painting in the style of Hokusai. (4) A photo of an Athenian vase with a painting of pandas playing basketball in the style of Egyptian hieroglyphics.}
  \label{fig:comparison-sota}
  \vspace{-1em}
\end{figure}

\paragraph{Training Setup}
Though multiple tasks are involved in the SFT process, a unified next-token-prediction loss is used for all of the tasks. As Lumina-mGPT is designed as a chat model, all data are organized into single or multi-turn dialogs, with the loss applied only to the response parts.
For all experiments, the AdamW~\citep{adamw} optimizer with weight decay = $0.1$ and betas = $(0.9, 0.95)$ is used, and the learning rate is set to 2e-5. Inspired by the classifier-free guidance in diffusion models~\citep{cfg}, we randomly drop the context by a probability of 10\% during training, as detailed in Appendix~\ref{app:infer}. To accommodate the large model volume, PyTorch FSDP~\citep{fsdp} is employed with gradient checkpointing. To increase training throughput, all data are pre-tokenized before training and are clustered according to the number of tokens, ensuring that each global batch is composed of data with similar lengths.

To stabilize training, we applied z-loss~\citep{palm} with a weight of \(1 \times 10^{-5}\) for both the 7B and 30B models. Additionally, dropout with a probability of 0.05 was applied specifically to the 7B model. Initially, we underestimated the significance of z-loss, as it is rarely mentioned in the literature on training multimodal large language models. However, our experiments revealed that omitting this term led to a surge in the magnitude of logits, resulting in diverging loss curves. Incorporating z-loss proved to be pivotal in enhancing training stability. Furthermore, we observed that applying z-loss significantly reduced the magnitude of logits, which in turn lowered the optimal temperature for inference-time image generation.

\begin{table*}[h]
\caption{Quantitative comparison on text-to-image benchmarks.}
\centering
\renewcommand{\arraystretch}{1.1}  
\setlength{\tabcolsep}{4.5pt} 
\resizebox{1.0\linewidth}{!}{%
\begin{tabular}{l cccc cccc ccc}
\toprule

\multirow{2}{*}{\textbf{Methods}} & \multicolumn{4}{c}{\textbf{GenEval} $\uparrow$} & \multicolumn{4}{c}{\textbf{DPG} $\uparrow$} & \multicolumn{3}{c}{\textbf{T2I-CompBench} $\uparrow$} \\ 
\cmidrule(lr){2-5} \cmidrule(lr){6-9} \cmidrule(lr){10-12}
& Two Obj. & Counting & Color Attri. & \textbf{Overall} & Entity & Relation & Attribute & \textbf{Overall} & Color & Shape & Texture \\ 
\midrule

\multicolumn{12}{l}{\textbf{Diffusion Models}} \\ \hline
SDv1.5~\citep{sd2_1}       & -  & -  & -  & 0.40  & 74.23  & 73.49  & 75.39  & 63.18  &0.3730 &0.3646 &0.4219      \\ 
Lumina-Next~\citep{lumina-next}  & 0.49  &  0.38  &  0.15  & 0.46  & 83.78      &89.78     & 82.67      & 75.66      &0.5088 &0.3386 &0.4239      \\ 
SDv2.1~\citep{sd2_1}     & 0.51  & 0.44  & 0.50  & 0.47  & -  & -  & -  & 68.09  &0.5694 &0.4495 &0.4982      \\ 
SDXL~\citep{sdxl}  & 0.74  & 0.39  & 0.23  & 0.55  & 82.43  & 86.76  & 80.91  & 74.65  &0.6369 &0.5408 &0.5637      \\ 
SD3-medium~\citep{sd3} & 0.74  & 0.63  & 0.36     & 0.62  &  91.01      &   80.70      & 88.83      & 84.08  & -      & -      & -      \\ 
DALL-E3~\citep{dalle3}   & 0.87    & 0.47     & 0.45  & 0.67 &89.61 & 90.58 &88.39 & 83.50 &0.8110&0.6750 &0.8070 \\ 

\midrule

\multicolumn{12}{l}{\textbf{AutoRegressive Models}} \\ \hline
LlamaGen~\citep{llamagen}     & 0.34  & 0.21  & 0.04  & 0.32  & -      & -      & -      & 65.16  & -      & -      & -      \\ 

Chameleon~\citep{chameleon}   & -     & -     & -     & 0.39  & -      & -      & -      & -      & -      & -      & -      \\ 

\rowcolor{green!10} 
Lumina-mGPT (Ours)&0.77 &0.27 &0.32 &0.56 &86.60 &91.29 &84.61 &79.68 &0.6371 &0.4727 &0.6034\\

\bottomrule
\end{tabular}%
\vspace{-1em}
}
\label{tab:t2i}
\end{table*}

\section{Experiments}
\label{sec:exp} 

\subsection{Photorealistic Text-to-Image Generation}
We first demonstrate the fundamental text-to-image generation capabilities achieved after FP-SFT. As shown in Figure~\ref{fig:demos}, Lumina-mGPT can generate photorealistic images in a variety of resolutions, achieving the first native 1K autoregressive generation without cascaded models~\citep{parti,imagen,muse,wurstchen}. Though finetuned on limited computational resources and text-image pairs, the results exhibit strong semantic coherence with intricate visual details

\paragraph{Qualitative Comparison}
Compared with LlamaGen~\citep{llamagen} and Parti~\citep{parti}, Lumina-mGPT achieves better visual quality and aesthetics (Fig.~\ref{fig:comparison-sota}). Besides, LlamaGen and Parti do not support end-to-end generation at high resolutions (e.g., 1k) or at arbitrary aspect ratios, whereas it is supported by Lumina-mGPT. Specifically, LlamaGen only supports a fixed resolution of  512 $\times$ 512, while Parti generates 1024 $\times$ 1024 images using an additional super-resolution upsampler. Beyond AR-based approaches, we also provide a side-by-side comparison with diffusion-based counterparts by training on the same dataset, as detailed in Appendix~\ref{app:diffusion}.
\reb{
\paragraph{Quantitative Comparison}
We evaluate the performance of Lumina-mGPT on widely used text-to-image benchmarks T2I-CompBench~\citep{huang2023t2icompbench}, GenEval~\citep{GenEval}, and DPG-Bench~\citep{hu2024ella}. The results are presented in Tab.~\ref{tab:t2i}.
}

Notably, Lumina-mGPT shows significant improvement over Chameleon~\citep{chameleon} and surpasses Lumina-Next~\citep{lumina-next}, a state-of-the-art diffusion transformer trained on the same text-to-image dataset as Lumina-mGPT. Moreover, Lumina-mGPT consistently outperforms SDv1.5 and SDv2.1~\citep{sd2_1}, becoming the first autoregressive model with competitive generation quality to SDXL~\citep{sdxl}. Nevertheless, a clear gap remains towards SOTA diffusion models like SD3~\citep{sd3} and DALL-E3~\citep{dalle3}, showing the room for further progress along this direction.
 
\begin{figure*}[t]
  \centering
  \includegraphics[width = \linewidth]{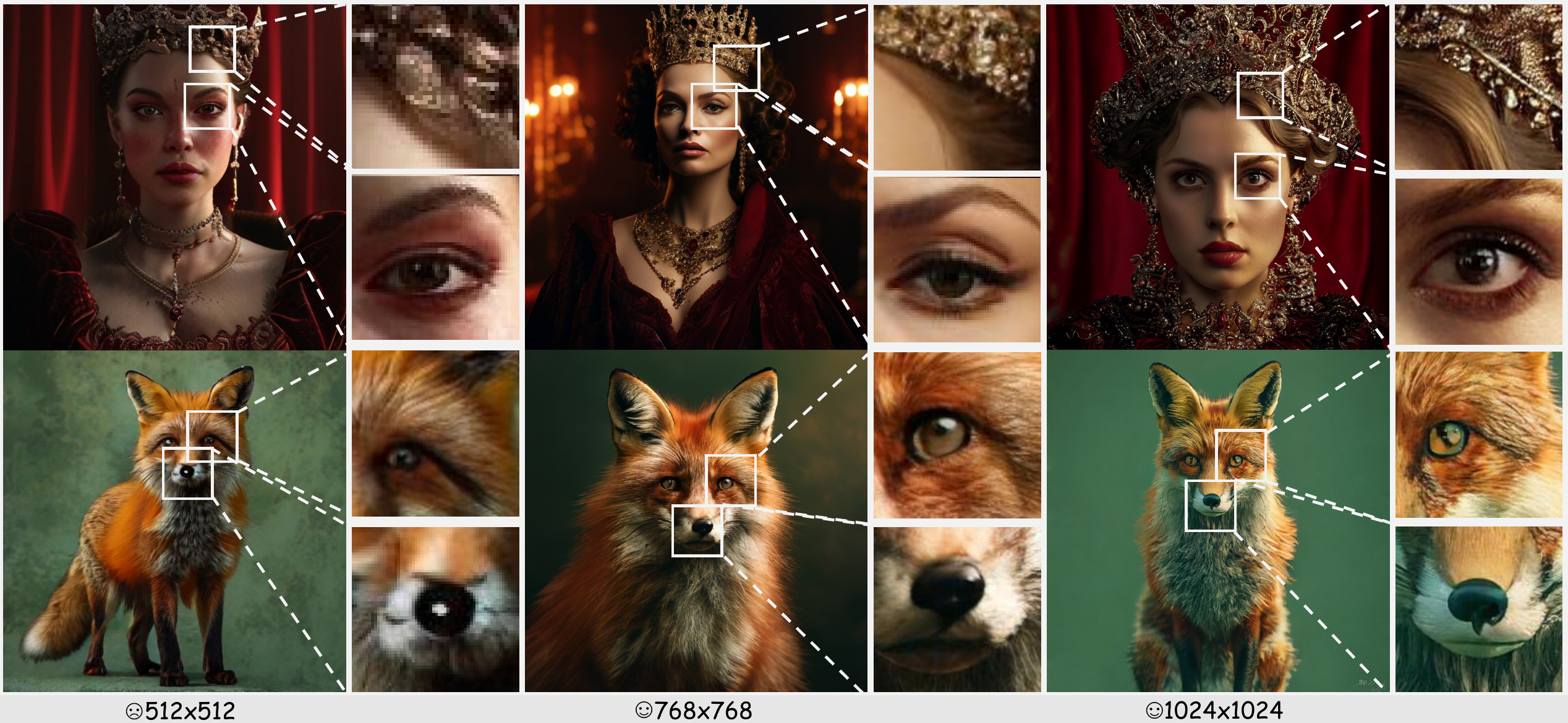}
  \vspace{-2em}
  \caption{Samples with zoom-ins generated by Lumina-mGPT in different resolution finetuning stages. The visual details continuously improve along with the progressively increasing resolution.}
  \label{fig:scaling_resolution}
\end{figure*}

\paragraph{On the Effectiveness of FP-SFT}
To further validate the effectiveness of FP-SFT, we visualize the images generated at different finetuning stages in Figure~\ref{fig:scaling_resolution}. With increasing image resolution, we observe a progressive decrease in visual artifacts introduced by VQ-VAE and the emergence of diverse fine-grained visual details. From these illustrations, we can conclude that our FP-SFT can unleash the potential of generating high-quality images from mGPT in a progressive manner.

\paragraph{Decoding Configuration Matters}
We observe the significant difference between the optimal hyperparameters for text and image generation, which inspires us to introduce a status-aware control mechanism switching settings based on the type of content being generated. We then explore how different inference configurations, such as temperature, top-k, and classifier-free guidance scale, affect the quality of generated images. For example, lower temperatures and top-k values often result in over-smoothed images, while higher values enhance detail but may introduce artifacts. Due to space limits, we defer the details to Appendix~\ref{app:infer}.

\begin{table}[t]
\centering
\vspace{-0.5em}
\caption{Evaluation on image understanding benchmarks.}
\small
\begin{tabular}{l|ccc}
\toprule
Method & MMBench & MME-p & MME-r \\
\midrule
Chameleon & 19.80 & 153.10 & 49.60 \\
Lumina-mGPT & \textbf{32.20} & \textbf{976.85} & \textbf{290.36} \\
\midrule
Method & SEEDBench-I & MMMU-val & POPE \\
\midrule
Chameleon & 30.50 & 22.40 & 19.40 \\
Lumina-mGPT & \textbf{50.93} & \textbf{27.11} & \textbf{70.43} \\
\bottomrule
\end{tabular}
\vspace{-1em}
\label{tab:vqa}
\end{table}

\subsection{Omnipotent Task Unification with Lumina-mGPT}
\label{sec:exp_omni_task_uni}

Beyond text-to-image generation, Lumina-mGPT also possesses the capability of text-only dialog and image understanding. After Omni-SFT, the spectrum of capabilities is further extended, covering multi-turn image editing, dense labeling, spatial-conditional image synthesis, and multiview generation. Such unification embodies the inherent benefits and potential of the decoder-only autoregressive architecture for unified modeling. Notably, the high-quality image generation ability achieved by Lumina-mGPT serves as the bedrock for such unification, without which the relevant explorations would be impossible.

To intuitively illustrate these capabilities, we provide visualizations in Figure~\ref{fig:demos_text} to~\ref{fig:demos_all_p4} in the Appendix.

First, Lumina-mGPT works effectively as a normal LLM to handle general text-only tasks  (Figure~\ref{fig:demos_text}), such as math problem solving, coding, and commonsense reasoning, thanks to the extensive pertaining in Chameleon and our involvement of text-only data to mitigate catastrophic forgetting.

Shown in Figure~\ref{fig:demos_all_p1}, Lumina-mGPT can also handle various visual understanding tasks including captioning, visual question answering, and general multi-turn multi-image dialog. Akin to image generation, flexible image aspect ratios are also supported in image understanding. Quantitative results on visual understanding benchmarks~\citep{mmbench,mmmu,pope,seedbench,mme} are presented in Table~\ref{tab:vqa}, illustrating that Lumina-mGPT significantly outperforms the original Chameleon model.

As a visual generalist, Lumina-mGPT incorporates classic visual recognition tasks. Using natural language as a unified interface, Lumina-mGPT can perform multiple high-level computer vision tasks including image segmentation, pose estimation, depth estimation, surface normal estimation, and referring object detection. See Figure~\ref{fig:demos_all_p2} and~\ref{fig:demos_all_p3} for examples.

Lumina-mGPT also supports image generation with versatile spatial conditions, such as depth maps, segmentation maps, normal maps, and human poses, as demonstrated in Figure~\ref{fig:demos_all_p3} and~\ref{fig:demos_all_p4}. Beyond single-image generation, Lumina-mGPT can produce a sequence of consistent images from multiview, as shown in Figure~\ref{fig:demos_all_p4}. This task marks a preliminary step toward video generation by rendering the multiview images into a coherent video. 

Given the above examples, though preliminary, they showcase that Lumina-mGPT can effectively follow diverse instructions, highlighting its promising potential as a unification of various challenging tasks in one framework. 

\begin{figure}[t]
  \centering
  \includegraphics[width = \linewidth]{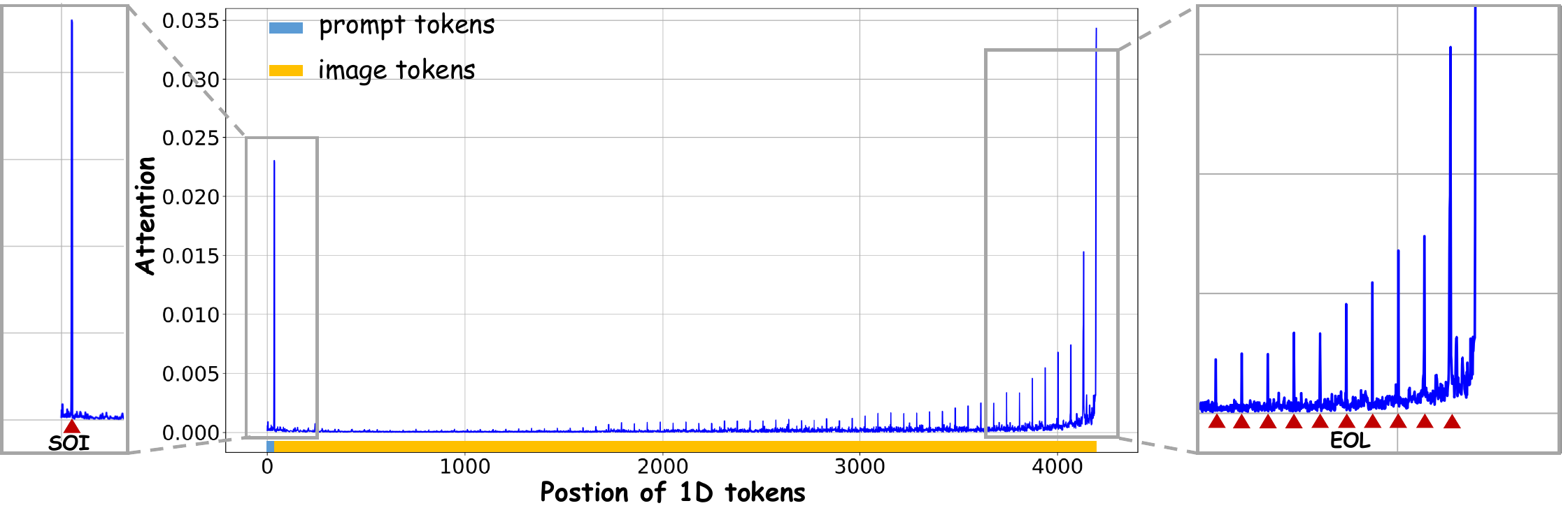}
  \vspace{-1.5em}
  \caption{Visualization of averaged attention logits from the last image token. Added indicator tokens are allocated with a large proportion of attention scores.}
  \label{fig:attention}
\end{figure}

\subsection{Attention Visualization}
\label{sec:attn_vis}
To gain deeper insights into the sampling behavior of Lumina-mGPT, we visualize the average attention logits of the final image token during text-to-image generation, as illustrated in Figure~\ref{fig:attention}. The results demonstrate that attention scores diminish for more distant tokens, indicating that the tokens attend more to local tokens over distant ones. This behavior is consistent with the local correlation nature of image data and also aligns with the long-term decay property inherent in the RoPE mechanism. Additionally, we observe that the indicator tokens, such as \texttt{<start-of-image>} and \texttt{<end-of-line>}, are assigned exceptionally high attention scores. This suggests that these indicator tokens play a pivotal role in the image-generation process. For instance, the high attention allocated to the \texttt{<start-of-image>} token implies that much of the semantic information from text tokens is encapsulated within it. Similarly, the high attention assigned to the \texttt{<end-of-line>} tokens indicates their effectiveness in introducing 2-D positional information to the 1-D RoPE positional embeddings.

\section{Conclusion}
\label{sec:conclusion}
In this work, we present Lumina-mGPT, a pioneering decoder-only multimodal autoregressive model that delivers flexible, high-quality photorealistic image generation from text prompts while unifying a wide range of vision and language tasks. Leveraging our novel \textit{Unambiguous Image Representation} (Uni-Rep) alongside the Flexible Progressive Supervised Finetuning (FP-SFT) and Omnipotent Supervised Finetuning (Omni-SFT) strategies, Lumina-mGPT achieves performance on par with modern diffusion models and surpasses existing autoregressive models like LlamaGen and Parti in visual quality and detail. Our extensive experiments demonstrate Lumina-mGPT's versatility as a multimodal generalist, effectively handling tasks from text-to-image synthesis to visual question answering. By releasing all code and checkpoints, we aim to foster further advancements and collaborations within the research community, paving the way toward more unified and powerful multimodal generative models.

\newpage
\bibliography{icml2025_submit}
\bibliographystyle{icml2025}

\newpage

\appendix
\onecolumn
\section{Additional Qualitative Results}
\begin{figure}[hb]
  \centering
  \includegraphics[width = 0.75\linewidth]{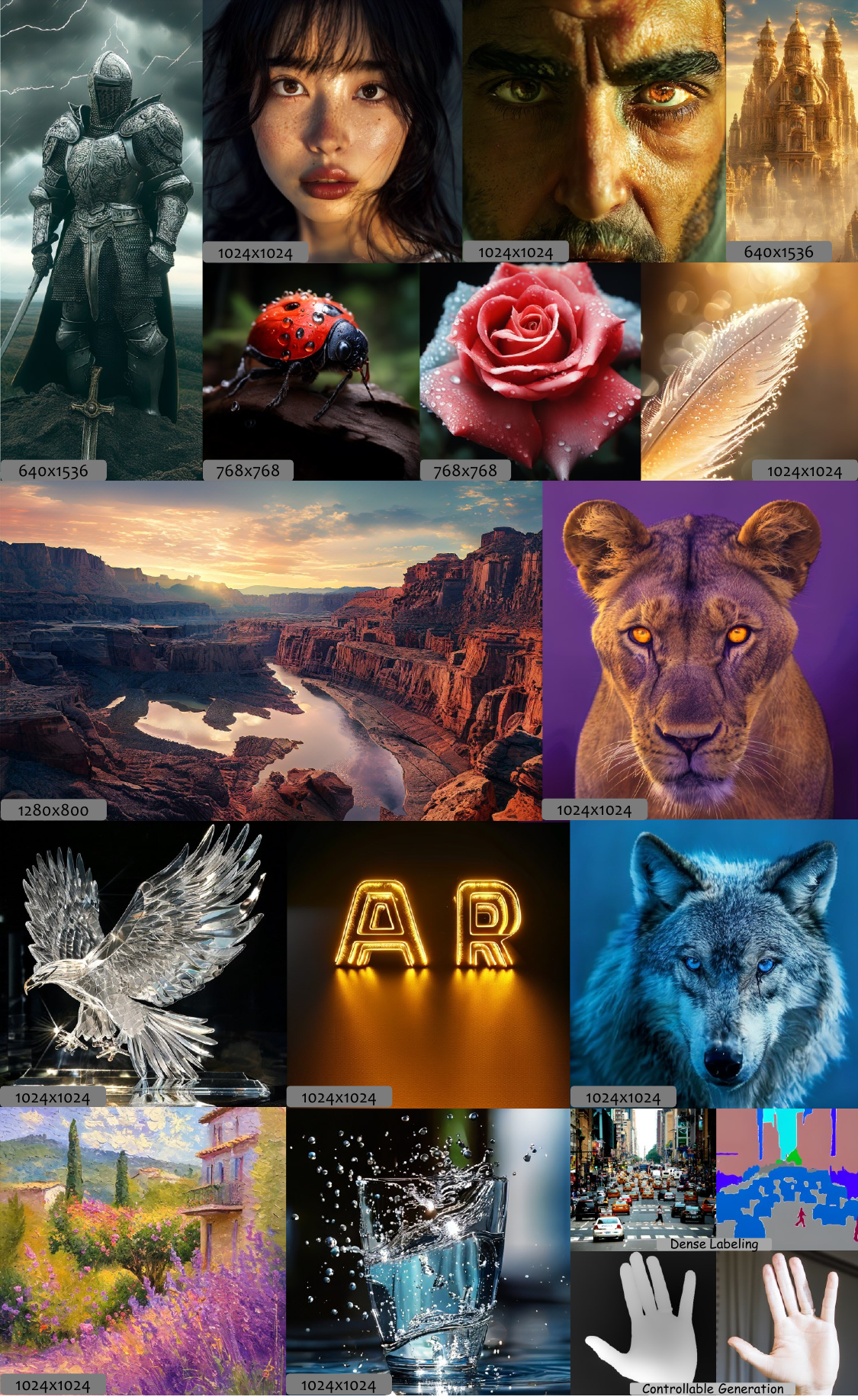}
  \caption{Selected photorealistic images of various resolutions generated by Lumina-mGPT. At the bottom, we include two examples to demonstrate the omnipotent task unification in Lumina-mGPT.}
  \vspace{-20em}
  \label{fig:demos}
\end{figure}

\begin{figure*}[htbp]
  \centering
  \includegraphics[width = 0.75\linewidth]{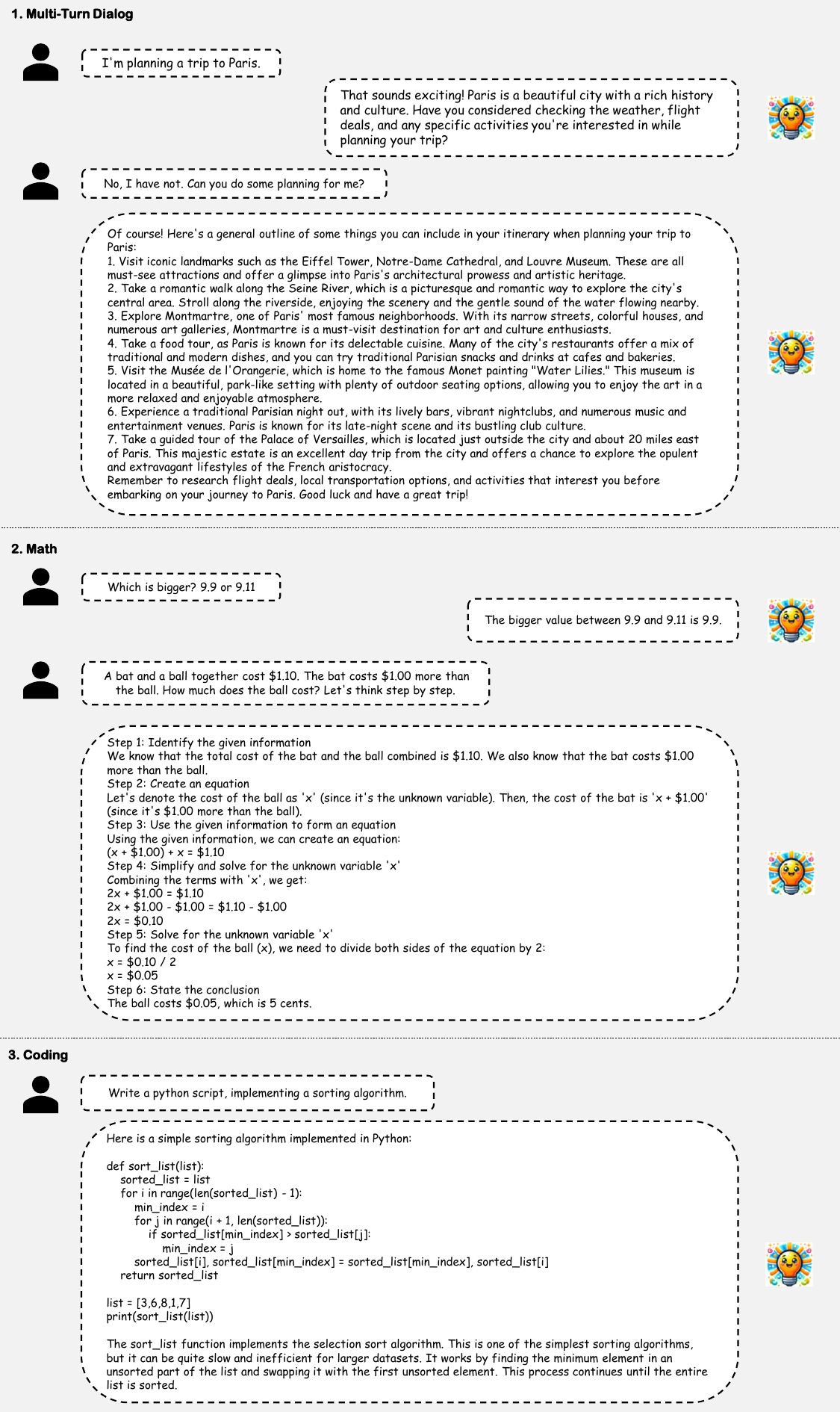}
  \caption{Lumina-mGPT as a generalist for various multimodal tasks.}
  \label{fig:demos_text}
\end{figure*}

\begin{figure*}[htbp]
  \centering
  \includegraphics[width = 0.8\linewidth]{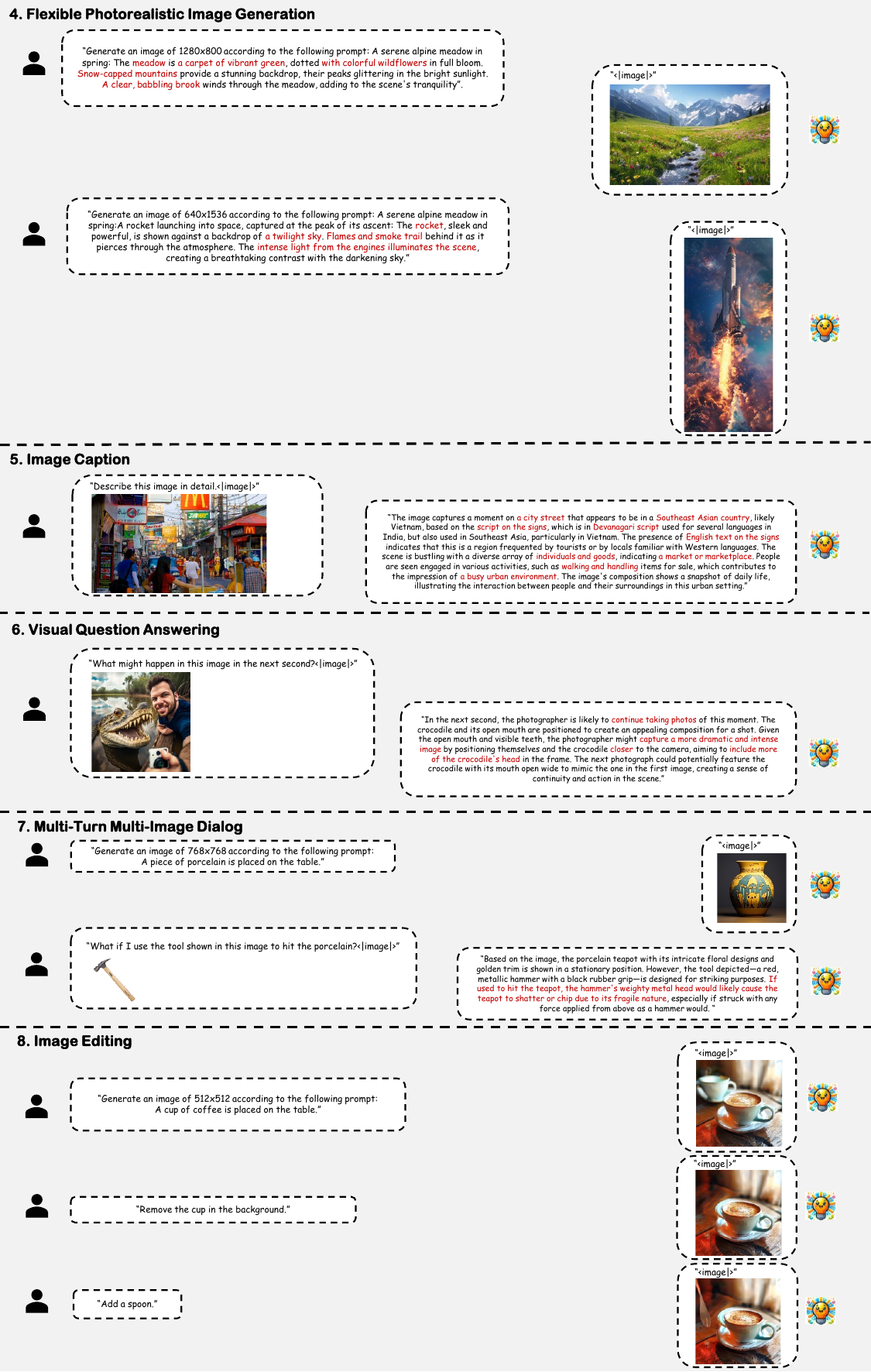}
  \caption{Lumina-mGPT as a generalist for various multimodal tasks.}
  \label{fig:demos_all_p1}
\end{figure*}

\begin{figure*}[htbp]
  \centering
  \includegraphics[width = 0.8\linewidth]{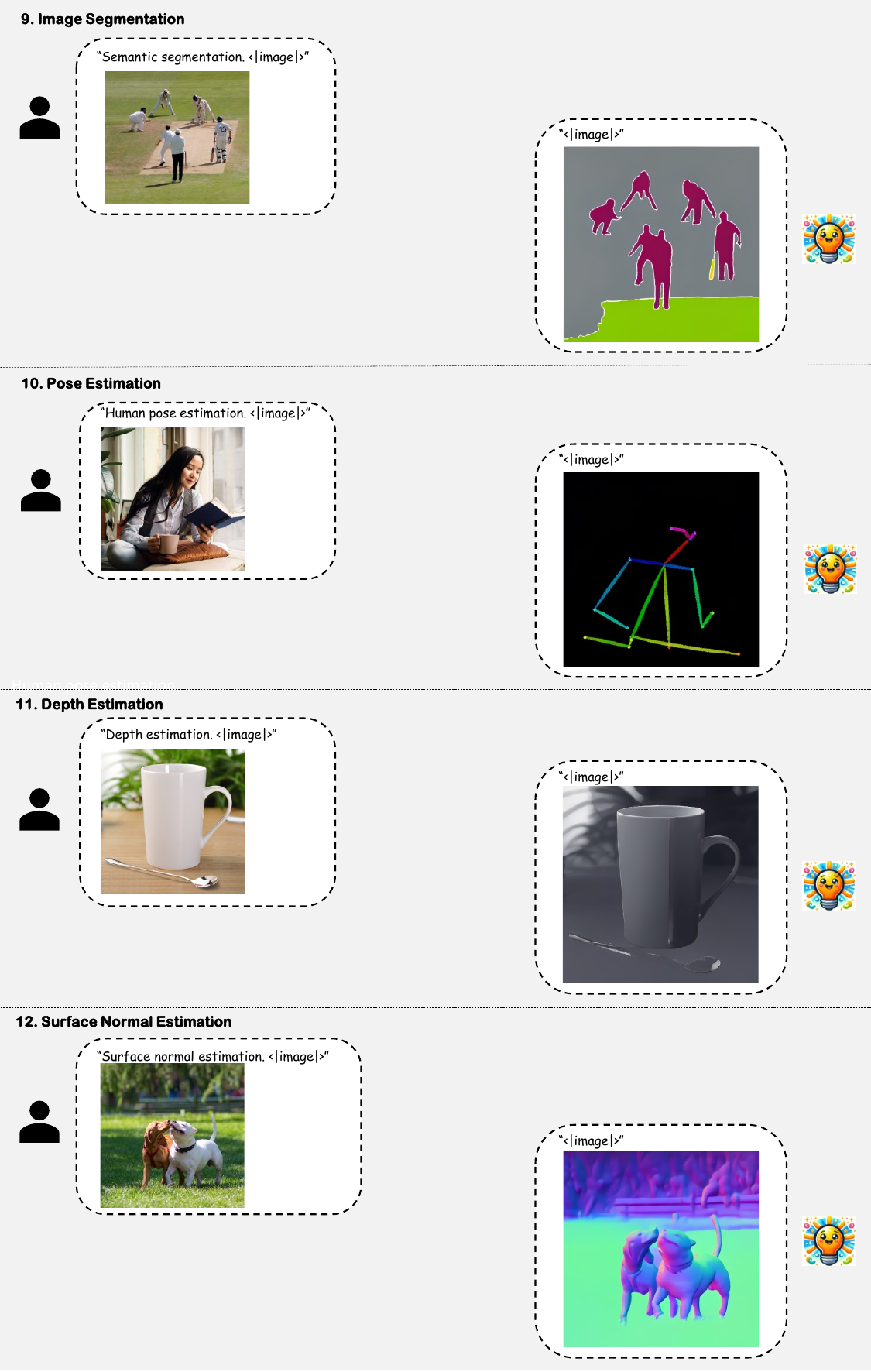}
  \caption{Lumina-mGPT as a generalist for various multimodal tasks.}
  \label{fig:demos_all_p2}
\end{figure*}

\begin{figure*}[htbp]
  \centering
  \includegraphics[width = 0.8\linewidth]{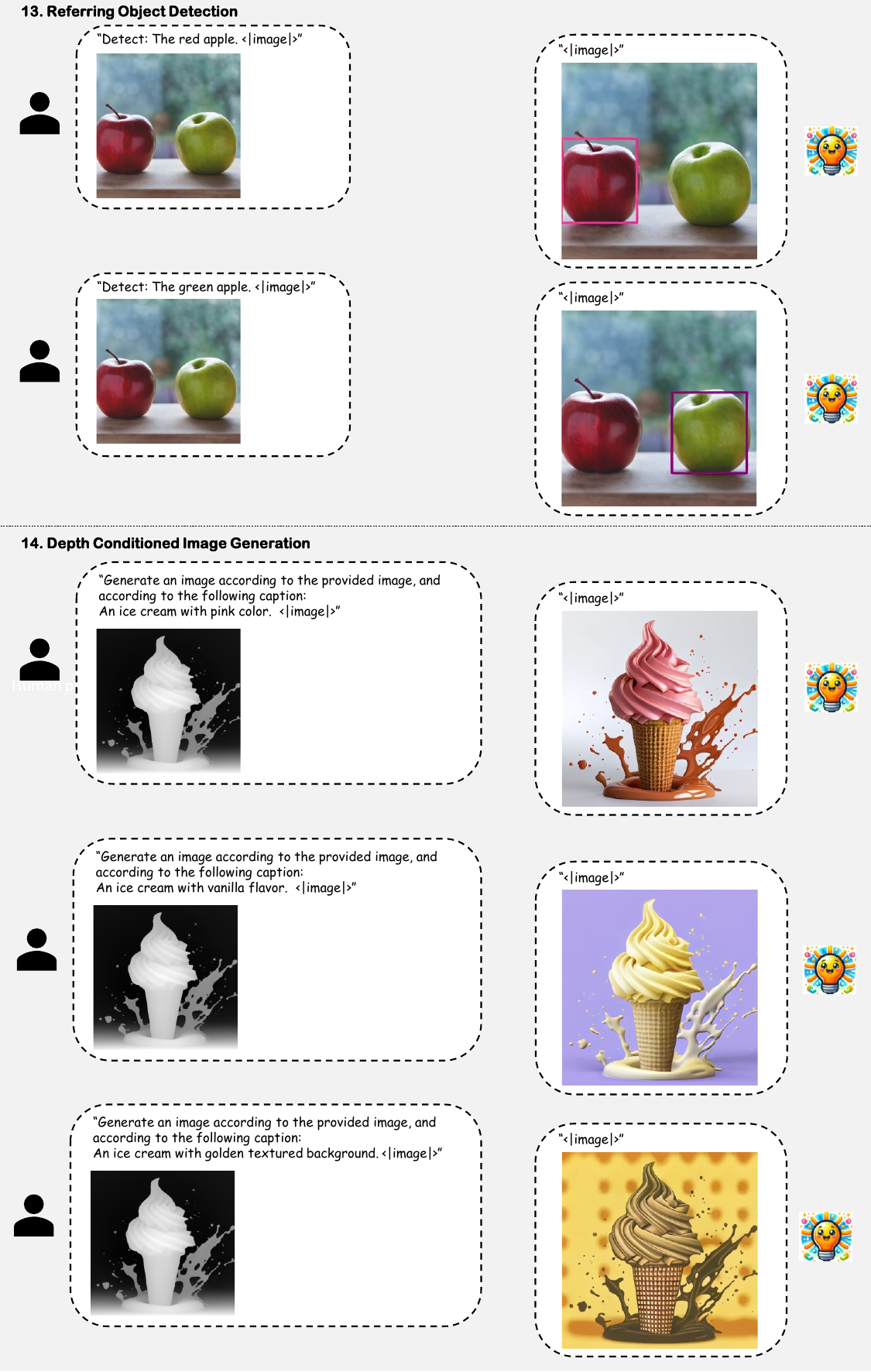}
  \caption{Lumina-mGPT as a generalist for various multimodal tasks.}
  \label{fig:demos_all_p3}
\end{figure*}

\begin{figure*}[htbp]
  \centering
  \includegraphics[width = 0.8\linewidth]{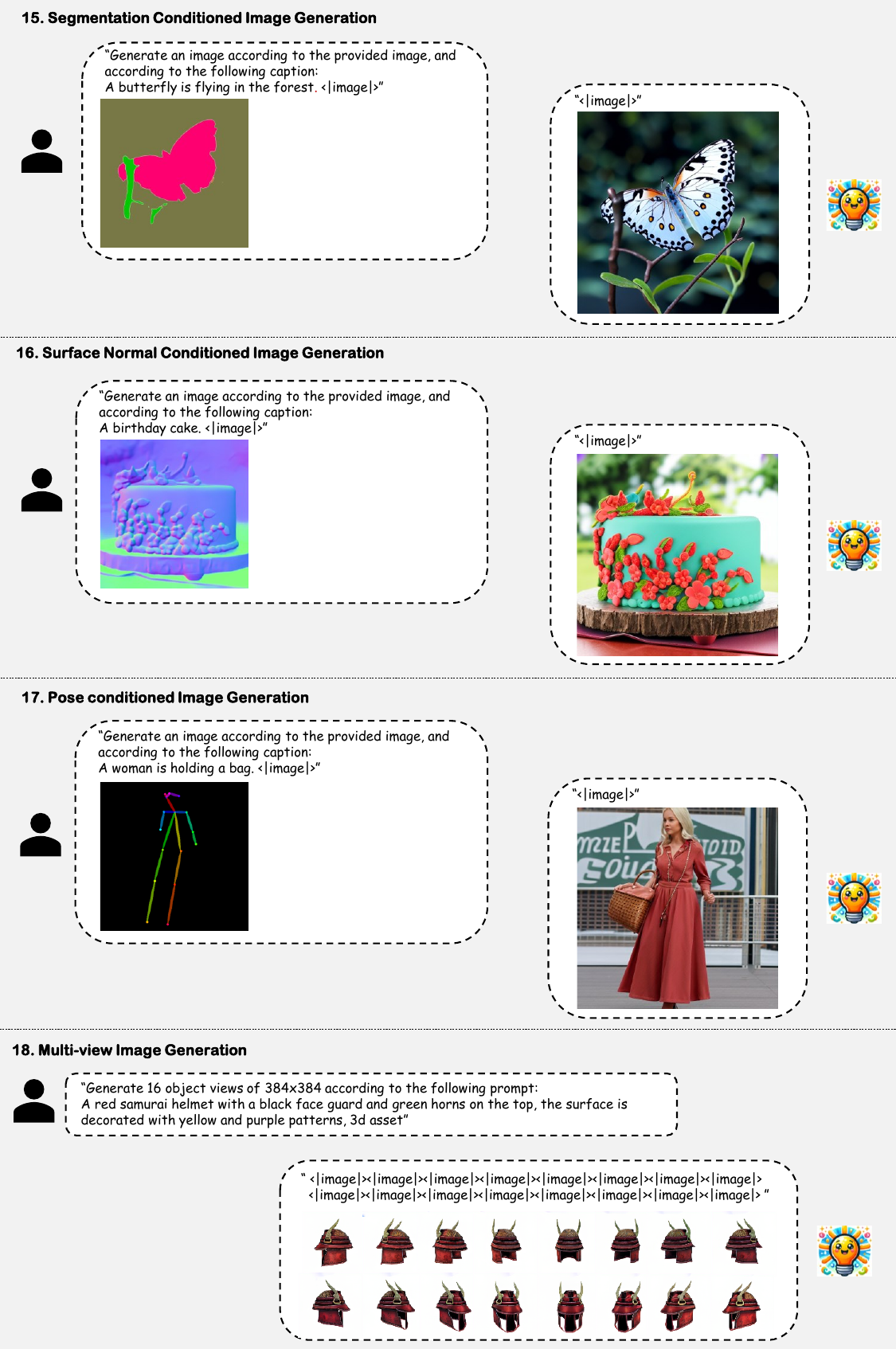}
  \caption{Lumina-mGPT as a generalist for various multimodal tasks.}
  \label{fig:demos_all_p4}
\end{figure*}

\clearpage

\twocolumn

\section{Related Work}
\label{app:related}

\paragraph{Multimodal Large Language Models}
Recent large language models (LLMs)~\citep{palm,llama,gpt3,gpt4,gemini} demonstrate strong instruction-following and reasoning capabilities, coupled with extensive world knowledge. To extend LLMs' expertise from the text domain to multimodal domains such as images and videos, prior works~\citep{llava,video-llava,video-chatgpt,sphinx} have aligned pretrained encoders from various modalities with LLMs by curating multimodal instruction tuning datasets. Although these multimodal large language models (MLLMs) have exhibited powerful visual understanding capabilities, their multimodal functionality is primarily limited to perceiving the visual world, exemplified by tasks such as visual question answering and image captioning, rather than generating visual outputs. 
Another line of research~\cite{gill,dreamllm,emu-llm,nextgpt} has proposed augmenting MLLMs with the ability to generate images, videos, and audio from text instructions. These approaches introduce additional visual tokens for generation and align these generative tokens as conditional information with a pretrained generator, such as Stable Diffusion~\citep{sd3,sdxl} for text-to-image generation. Consequently, the generation capabilities heavily rely on the external expert generator rather than MLLMs themselves, resulting in inconsistent and inferior generation results. To combine the strength of both approaches, our model aims to learn both understanding and generation of images using an MLLM with native multimodal capabilities, drawing inspiration from Chameleon~\citep{chameleon}, a mixed-modal early-fusion foundation model. 

\paragraph{Text-to-Image Generation}
The task of text-to-image generation seeks to synthesize photorealistic and diverse images based on textual descriptions. Nowadays, diffusion models, whether in pixel space~\citep{imagen} or in latent space~\citep{sdxl,sd3,wurstchen}, have become the de-facto approaches due to their superior performance, particularly in producing extremely high-aesthetic images. Among these models, the recent trend of scaling diffusion transformers (DiTs)~\citep{pixart-alpha,pixart-sigma,lumina-t2x,sd3,hunyuandit,kolors} suggests a unified architecture for both text and image modeling. However, existing DiTs still leverage separate language models, such as CLIP~\citep{clip} or T5~\citep{t5}, as text encoders. This modality gap between text and image representations not only leads to inaccurate generation but also hinders the development of a unified multimodal foundational generative model.
Compared to the dominance of diffusion models, the progress of autoregressive image generation has received less attention in recent years. Early works~\citep{dalle,cogview} proposed a two-stage generation approach: first, training a VQ-VAE~\cite{vq-vae,vqgan} for image tokenization and de-tokenization, and then using an autoregressive transformer to model discrete image token sequences, akin to language modeling. Parti~\citep{parti} scaled up the autoregressive transformer to 20 billion parameters, demonstrating promising high-fidelity image generation results. LlamaGen~\citep{llamagen} further improved the image tokenizer and integrated advanced techniques in LLMs, bridging the performance gap with diffusion counterparts. Unlike Parti and LlamaGen, Lumina-mGPT proposes multimodal generative pertaining on unified text-image sequences, followed by supervised finetuning on high-quality text-to-image pairs, achieving flexible high-aesthetic image generation with autoregressive models.

\section{\reb{Discussions on Reconstruction Quality}}

\begin{figure}[t]
  \centering
  \includegraphics[width = \linewidth]{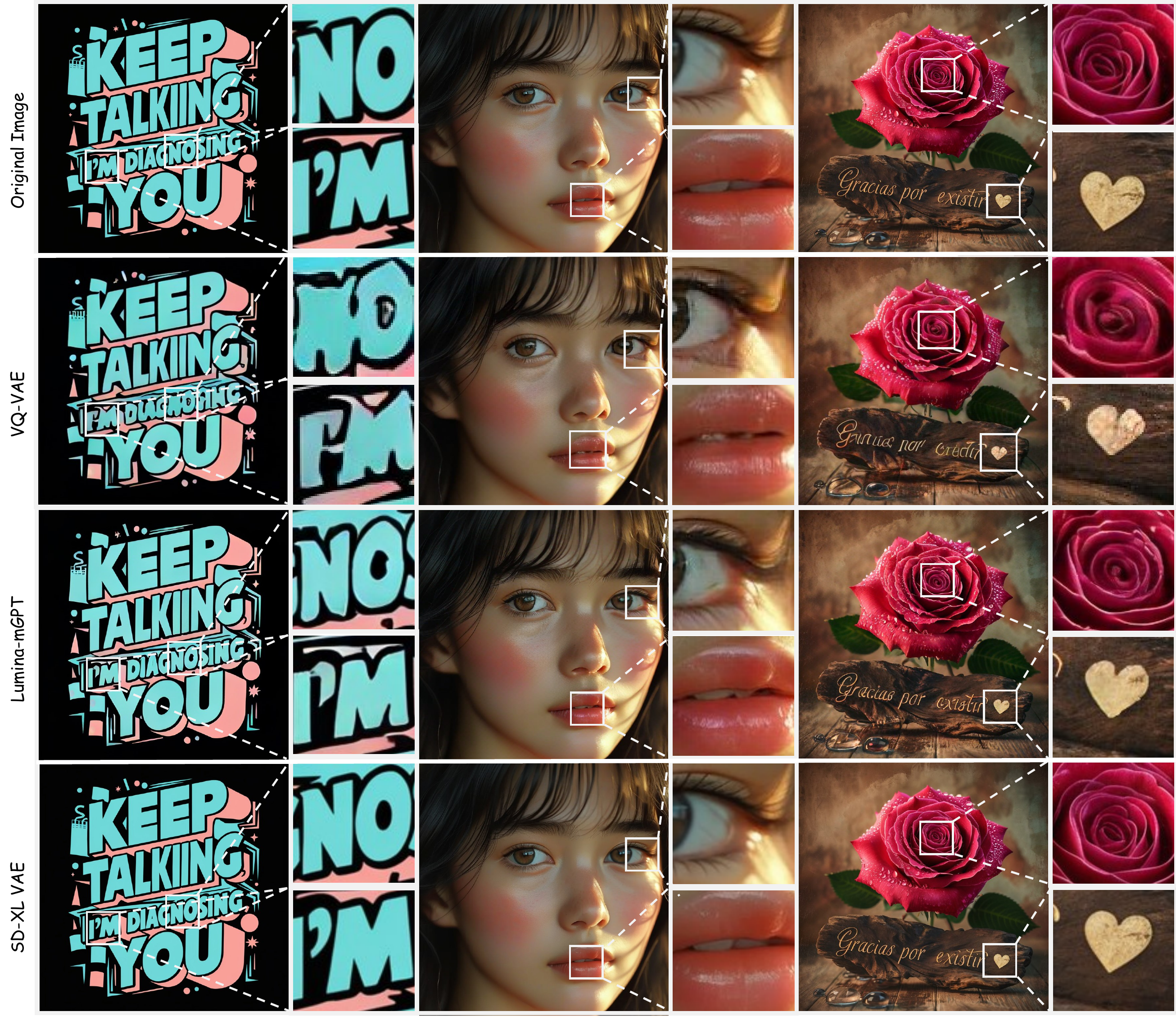}
  \vspace{-4mm}
  \caption{\reb{Reconstruction quality of different methods. Lumina-mGPT means first encode using VQ-VAE encoder, then send the latent to Lumina-mGPT using image editing system prompt with the instruction "no edit", and finally decode the newly generated latents using VQ-VAE decoder.}}
  \label{fig:reconstruct}
  \vspace{-4mm}
\end{figure}

\reb{
VQ-VAEs~\citep{vq-vae,vavqe2,vqgan} compress images at the cost of information loss, which introduces quality degradation at reconstruction, especially for high-frequency details such as edges, hair, and text. As Generative models such as Lumina-mGPT only has access to the VQ-VAE latents during training and cannot access original images, intuitively the VQ-VAE reconstruction quality should somehow build an upper bound for such models' image generation quality.}

\reb{
However, we observe an interesting and counter-intuitive phenomenon. Given two data flows:
}

\small{\reb{
1. Image $\xrightarrow{\text{VQVAE Encoder}}$ latent$\xrightarrow{\text{VQVAE Decoder}}$ Recon1
}}

\small{\reb{
2. Image $\xrightarrow{\text{VQVAE Encoder}}$ latent$\xrightarrow{\text{Lumina-mGPT using editing system prompt with instruction "no edit"}}$ latent2$\xrightarrow{\text{VQVAE Decoder}}$ Recon2
}}

\reb{
We surprisingly find that the quality of Recon2 sometimes surpasses that of Recon1, and we show such cases in Fig.~\ref{fig:reconstruct}. For reference, the reconstruction results using SDXL VAE~\citep{sdxl} are also presented. Note that while Lumina-mGPT has been trained on the image editing task, it is not trained with the "no edit" instruction. This intriguing observation may suggest some meaningful insights. For example, it may possibly indicate that the latents encoded by VQ-VAE encoder could contain noises of certain patterns that can be learned and even corrected. We leave the further exploration of this phenomenon for future work.
}

\section{Limitations of Existing Approaches}
\label{app:limit_exist}
\textbf{Randomly-Initialized Transformer} While transfer learning has revolutionized key fields such as visual recognition~\citep{resnet,clip,vilbert} and language generation~\citep{t5,gpt1,gpt2,gpt3}, popular autoregressive image generation approaches such as DALL-E, Parti, and LlamaGen all adopt a randomly-initialized causal transformer, which fails to utilize pretrained transferable representation and large-scale datasets. As a result, AR-based approaches often lead to poor image generation quality and slow convergence without leveraging proper large-scale pretraining. 

\textbf{Verbose Encoder-Decoder Architeture} DALL-E and CogView initially propose using a decoder-only transformer for image generation with discrete representation, where a single transformer acts as both a text encoder and an image token decoder. However, subsequent approaches, such as Parti and LlamaGen, adopt a verbose encoder-decoder architecture that injects frozen T5 text features~\citep{t5} using cross-attention or prefix-filling approaches, motivated by the findings of Imagen~\citep{imagen}. Compared to the trend in scaling LLMs~\citep{llama,qwen,deepseek}, such encoder-decoder architecture is cumbersome due to the decoupling of text encoding and image token modeling. This design significantly complicates the autoregressive-decoding framework, limits the scalability of image generation, and hinders the generalization to additional modalities and tasks.

\textbf{Limited Decoding Resolution and Flexibility} Natural images exsit in various resolutions and aspect ratios. Advanced diffusion models~\citep{pixart-sigma,pixart-alpha,sd3,kolors,hunyuandit,lumina-next,fit} can successfully generate diverse photorealistic images at arbitrary resolution with skewed ratios. In contrast, current AR-based approaches~\citep{parti,chameleon,llamagen} rely on central-cropping a low-resolution 512 $\times$ 512 image and transforming the cropped low-resolution image into a fixed-length sequence of discrete tokens using a pretrained Vector-Quantized Variational Autoencoder (VQ-VAE)~\citep{vq-vae,vavqe2,vqgan}. This approach simplifies autoregressive training but at the cost of deteriorated image quality and generation flexibility. 

\textbf{Poor Task Extensability} Autoregressive modeling excels at unified generative modeling for various tasks and modalities using discrete tokens~\citep{ofa,unified-io,unified-io2}. However, previous AR-based image generation approaches~\citep{dalle,cogview,parti,llamagen} have been limited to text-to-image generation without exploring the unification with other tasks, such as dense labeling and controllable image generation. This lack of task extensibility significantly constrains the applicability of AR-based image generation models to a broader range of scenarios.

\section{Limitations of Lumina-mGPT}
\label{sec:limitation}

\begin{figure}[t]
  \centering
  \includegraphics[width = \linewidth]{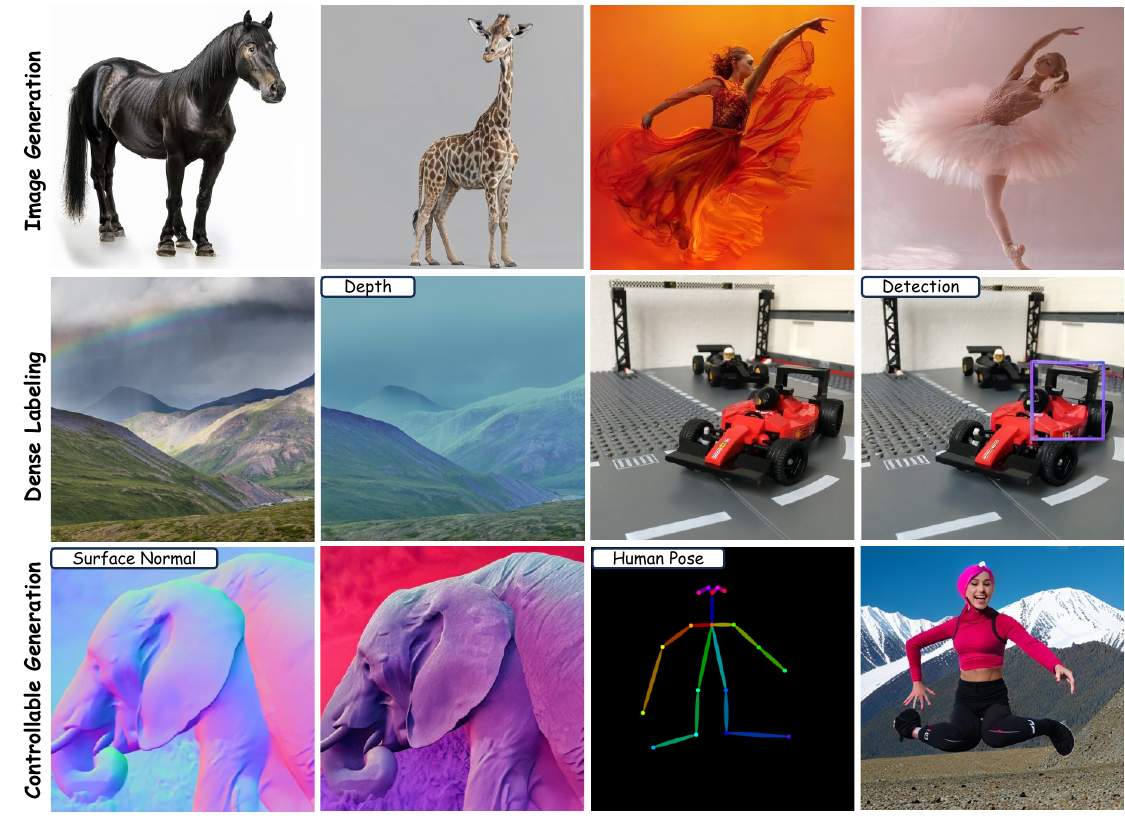}
  \caption{Failure cases of current Lumina-mGPT. Due to inadequate training and limited data size, Lumina-mGPT sometimes struggles to understand input conditions and produce visual artifacts.}
  \label{fig:failure}
\end{figure}

\paragraph{Failure Cases}
Despite Lumina-mGPT can generate photorealistic images, it sometimes produces images with noticeable visual artifacts. For example, Lumina-mGPT may generate people and animals with unreasonable limbs, as shown in the first row of Figure~\ref{fig:failure}. Besides, compared to SoTA text-to-image 
generation approaches including SD3~\citep{sd3}, Kolors~\citep{kolors}, and HunyuanDiT~\citep{hunyuandit}, all pretrained over 1B image-text pairs, Lumina-mGPT's prompt-following ability is inferior due to the limited training resources and data size, which are many times smaller than these SoTA methods.
Regarding dense labeling and controllable generation, Lumina-mGPT currently showcases preliminary results with a limited training budget. Hence, the second row in Figure~\ref{fig:failure} provides such an example where Lumina-mGPT produces inaccurate predictions or semantically inconsistent images, failing to understand the given image conditions. 
Therefore, we expect by scaling data size with more computational resources, Lumina-mGPT can effectively address the above failure cases such as inadequate instruction-following ability and visual artifacts.

\paragraph{Generation Speed}
Autoregressive models require numerous network evaluations during inference due to the nature of next-token prediction, similar to the iterative denoising process in diffusion models. This becomes worse when generating high-resolution images, which often require minutes to generate a full sequence of image tokens, significantly slower than current diffusion models with advanced samplers. However, there have been plenty of techniques to optimize the inference speed designed for autoregressive models, such as vLLM~\citep{vllm} and FlashAttention~\citep{flashattention,flashattention2}. We believe that by integrating these approaches in the future, Lumina-mGPT can achieve a remarkable speed up during inference.

\paragraph{VQ-VAE Reconstruction Quality}
VQ-VAE is employed as the image tokenizer to convert continuous images into discrete token representations. Meanwhile, it also introduces information bottlenecks by compressing the spatial dimensions of images. As a result, the reconstruction quality of VQ-VAE largely determines the upper limit of generation quality. We discover that the VQ-VAE proposed in Chameleon sometimes struggles to reconstruct high-frequency details, especially when text and human faces are present in images. Incorporating further improvements on VQ-VAE, such as FSQ~\citep{fsq}, may also enhance the generation quality of Lumina-mGPT.

\begin{figure}[t]
  \centering
  \includegraphics[width = \linewidth]{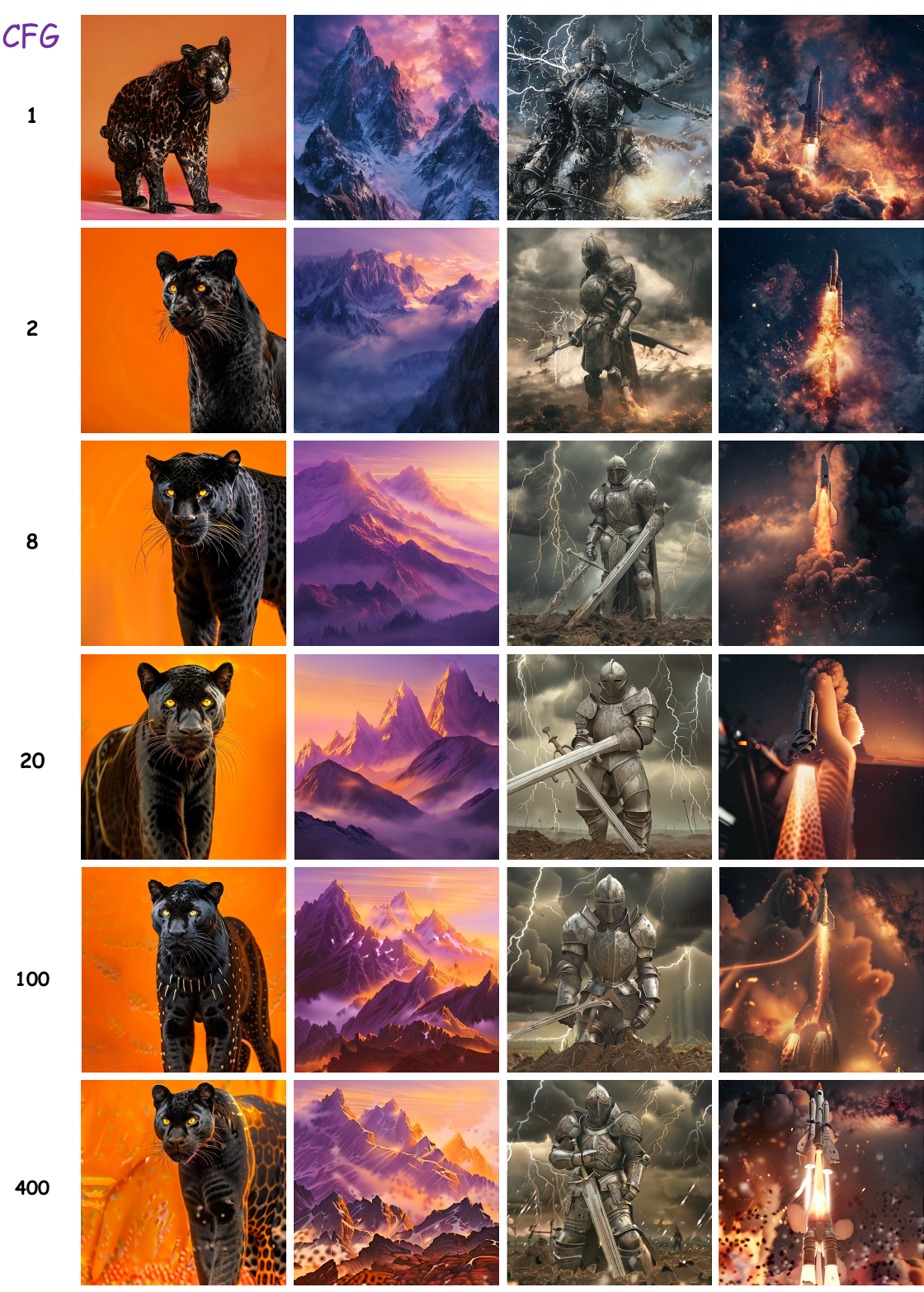}
  \vspace{-6mm}
  \caption{\reb{Samples generated by Lumina-mGPT using different CFG; T=1.0, Top-k=2000.}}
  \vspace{-2mm}
  \label{fig:decoding-config-cfg}
\end{figure}
\section{Inference Configuration of Lumina-mGPT}
\label{app:infer}

In autoregressive models, various configuration parameters during Lumina-mGPT's decoding stage significantly affect sample quality~\citep{Holtzman2020The,gpt1,gpt2}. Hyperparameters such as temperature (T), top-k, and classifier-free guidance scale (CFG) have not been extensively investigated in the visual domain. In this section, we explore how these hyperparameters influence the generated image in terms of quality, texture, and style.

\paragraph{Different Decoding Hyperparameters for Image and Text}
The sampling strategy of autoregressive models involves numerous hyperparameters that significantly influence the sampling results. We find that the optimal decoding hyperparameters differ greatly between text decoding and discrete image code decoding. For example, the top-k=5 setting performs well in generating text. However, when generating images, the value of top-k should be much larger (e.g. 2000) to avoid repetitive and meaningless patterns. Therefore, we implement a status-aware control mechanism for inference. Specifically, a set of default hyperparameters is used for text decoding; once a \texttt{<start-of-image>} token is generated, the hyperparameters switch to those optimized for image generation. After the \texttt{<end-of-image>} token is generated, the parameters revert to the initial settings.

\paragraph{Classfier-Free Guidance} Classifier-Free Guidance (CFG) \citep{cfg} is originally proposed to enhance the quality and text alignment of generated samples in text-to-image diffusion models. We incorporate this technique into autoregressive models during inference. When generating an image token, the CFG-processed logits \( l_{\text{cfg}} \) are formulated as \( l_{\text{cfg}} = l + s(l - l') \), where \( l \) represents the original logits conditioned on the complete context; \( l' \) represents the context-independent logits, which are conditioned solely on the tokens following the \texttt{<start-of-image>} token of the currently generating image, and are independent of any prior context; $s$ denotes the guidance scale of Classifier-Free Guidance. To make CFG work, during training, the context before <start-of-image> is randomly dropped by a probability of 10\%. In practice, KV cache can be used for accelerating the computation of both $l$ and $l'$. \reb{As shown in Fig.~\ref{fig:decoding-config-cfg}, similar to the trend of diffusion models, increasing CFG initially raises the quality and stability of generation, but increasing it further would make the quality deteriorate.}

\begin{figure}[t]
  \centering
  \includegraphics[width = \linewidth]{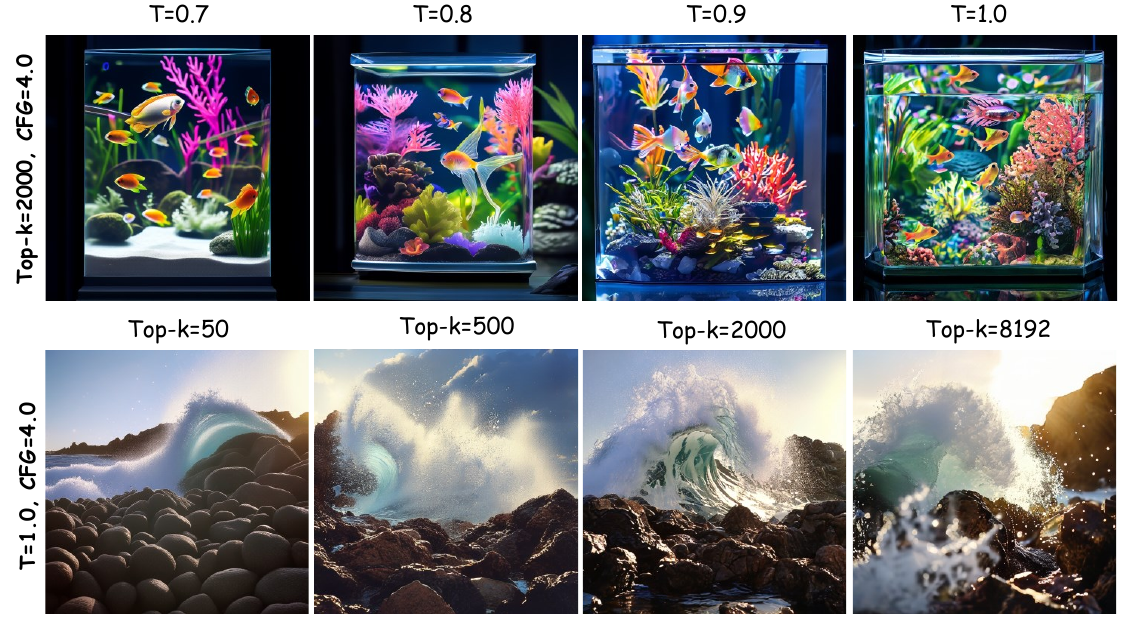}
  \vspace{-2mm}
  \caption{Samples generated by Lumina-mGPT using different Top-k and Temperature.}
  \vspace{-4mm}
  \label{fig:decoding-config}
\end{figure}

\paragraph{Influence of Temperature}
To evaluate the effect of these decoding parameters, we first set a standard decoding configuration: T=1.0, top-k=2000, CFG=4.0, which serves as a good-to-use setting. From this baseline, we gradually shift T from 0.7 to 1.0 to generate corresponding images at different temperatures. As shown in Figure~\ref{fig:decoding-config}, it is evident that when setting the temperature low, visual details diminish and objects tend to be over-smoothed. Conversely, when setting the temperature high, the generated images contain rich visual content but are prone to contain more artifacts.

\paragraph{Influence of Top-k}
Based on the standard decoding setting, we vary the top-k value, from 50 to 8192, where 8192 is equal to the size of the VQ-VAE codebook used usedin Lumina-mGPT. The results, visualized in Figure~\ref{fig:decoding-config}, indicate a similar trend with increasing temperature. When top-k is low, the image content and texture are relatively simple, exhibiting the over-smoothed problem as well. When top-k is set high, the image detail and texture are diverse, making it more aesthetically appealing, while increasing the potential of artifacts.

\begin{figure}[t]
  \centering
  \includegraphics[width = \linewidth]{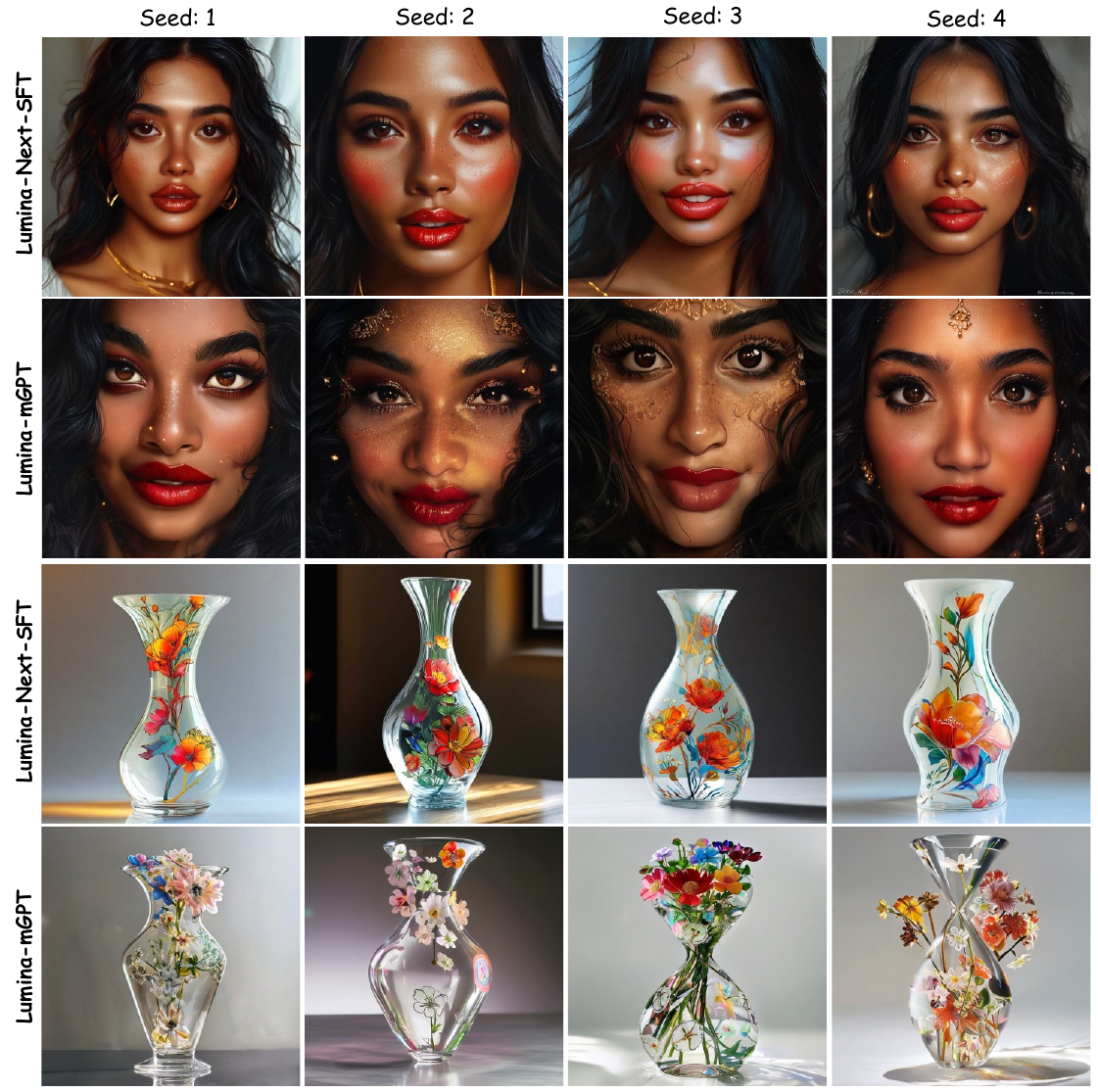}
  \caption{Comparison between Lumina-Next-SFT and Lumina-mGPT using different random seeds. Images generated by Lumina-mGPT exhibit comparable aesthetics with greater diversity.}
  \label{fig:comparison}
\end{figure}

\section{Comparison with Diffusion-based Appraoches}
\label{app:diffusion}
For a long period of time, diffusion models have dominated the field of text-to-image generation compared to autoregressive models. Although LlamaGen claims to beat diffusion models, their results are limited to the ImageNet benchmark and there has been no direct comparison between these two architectures so far. In this section, we aim to provide a detailed comparison of autoregressive and diffusion-based methods trained on these same text-image datasets, focusing on image quality, diversity, text-rendering, and multilingual capabilities. Specifically, we adopt Lumina-mGPT and Lumina-Next-SFT~\citep{lumina-next} as representatives of autoregressive and diffusion-based methods, respectively. A direct visual comparison between Lumina-Next-SFT and Lumina-mGPT reveals both the similarities and differences between autoregressive and diffusion-based generative modeling approaches. 

\begin{figure}[t]
  \centering
  \includegraphics[width = \linewidth]{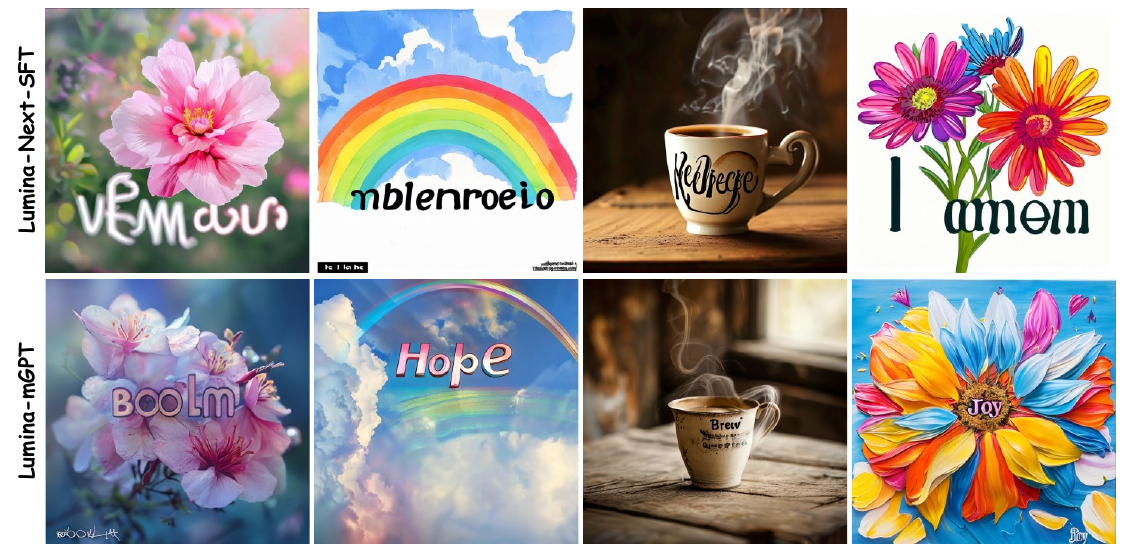}
  \caption{Text rendering comparison between Lumina-Next-SFT and Lumina-mGPT. From left to right, the correct texts to be rendered on the image are: ``Bloom'', ``Hope'', ``Brew'', and ``Joy''.}
  \label{fig:comparison-text}
\end{figure}

\begin{figure}[t]
  \centering
  \includegraphics[width = \linewidth]{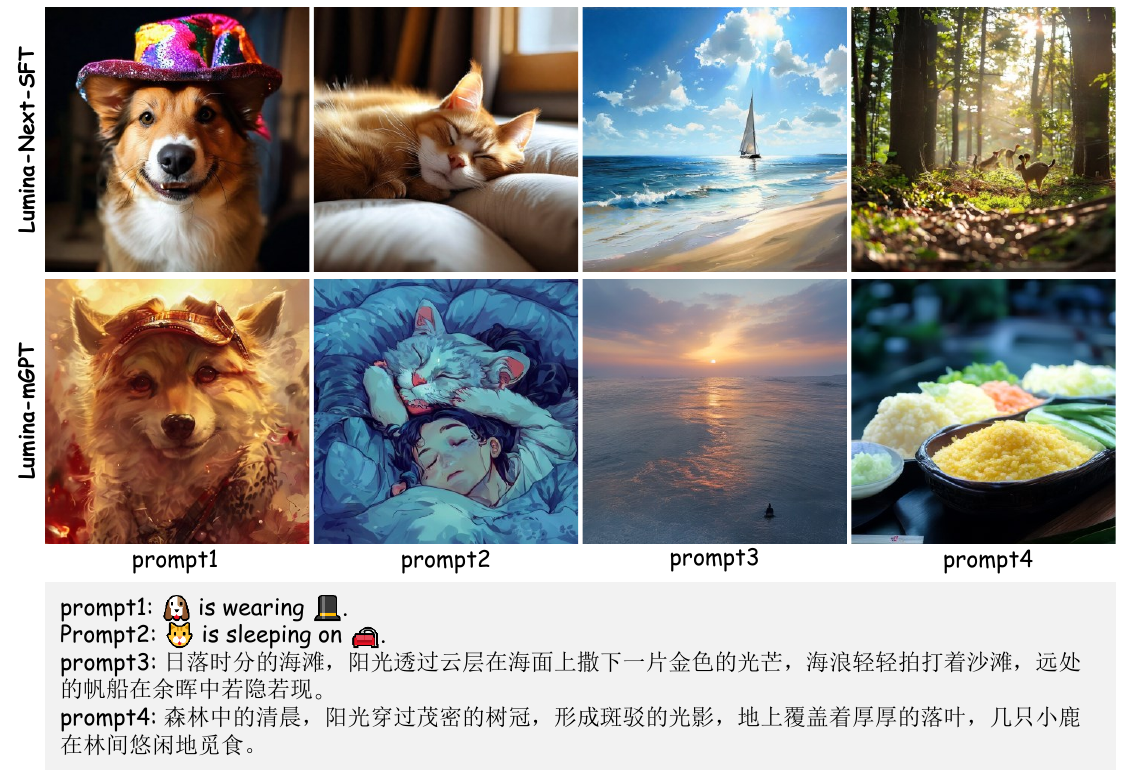}
  \caption{Emoji and multilingual instruction understanding ability comparison between Lumina-Next-SFT and Lumina-mGPT. Lumina-mGPT struggle to understand emojis and multilingual prompts.}
  \label{fig:comparison-multilingual}
\end{figure}

\textbf{On the Similarity between Diffusion- and AR-based Generation}
Given the same set of text prompts, both diffusion- and AR-based approaches generate photorealistic images with similar aesthetic style and fine-grained details, illustrated in Figure~\ref{fig:comparison}. This reveals the fact that both architectures can achieve satisfactory text-to-image generation performance when provided with the same training data, training budget, and comparable model sizes. The AR-based methods display remarkable visual aesthetics on par with their diffusion counterparts, challenging the notion that diffusion models are more effective and promising architecture in generative modeling. This finding also aligns with the platonic representation hypothesis~\citep{huh2024prh} that neural networks are converging to learn a shared representation space despite being trained with different architectures and objectives. Therefore, this hypothesis highlights the importance of collecting more high-quality data and optimizing training infrastructure as directions for data and model scaling, to boost the overall model performance that is agnostic to any specific architecture.

\textbf{On the Differences between Diffusion- and AR-based Generation}
As shown in Figure~\ref{fig:comparison}, Lumina-mGPT exhibits more diversity using different random seeds, while Lumina-Next-SFT generates similar images with identical layouts and textures. This can be partly attributed to the use of high temperature and top-k values in Lumina-mGPT. However, excessive diversity also causes our model to be less stable and more prone to producing visual artifacts, which is discussed in Section~\ref{sec:limitation}. 

We also compare the text rendering and multilingual understanding capabilities between Lumina-mGPT and Lumina-Next-SFT. As illustrated in Figure~\ref{fig:comparison-text}, Lumina-mGPT exhibits significantly better text synthesizing results, while Lumina-Next-SFT struggles to generate any complete character. We argue that this underscores the importance of mGPT, where the model learns a seamless multimodal representation between text and images using massive interleaved data during the pertaining stage. However, when it comes to multilingual understanding, Lumina-mGPT performs worse than Lumina-Next-SFT in terms of emoji and Chinese prompts shown in Figure~\ref{fig:comparison-multilingual}. The reason is that although Lumina-mGPT learns better text-image alignment, the lack of multilingual text corpus used in pertaining limits its performance. In contrast, the text encoder used in Lumina-Next-SFT showcases significantly stronger multilingual capabilities than Chameleon. Hence, we hope that by comprehensively enhancing the capabilities of the base mGPT model, such as adding more multilingual data, Lumina-mGPT can benefit in all downstream tasks.

In addition to text-to-image generation, Lumina-mGPT supports various vision and language tasks within a unified framework. However, the design of diffusion models limits their compatibility and performance across multiple modalities and tasks. They often require specific architecture designs and additional training for each unseen task~\citep{marigold,odise}. In contrast, Lumina-mGPT treats input from all modalities as multimodal token sequences and leverages natural language as the interface to unify diverse tasks with next-token prediction.

\bibliography{icml2025_submit}
\bibliographystyle{icml2025}

\end{document}